%% file: template.tex
\documentclass[twocolumn]{svjour3}          
\smartqed  
\usepackage{graphicx}
\usepackage{epsfig}
\usepackage{amsmath}
\usepackage{mathtools}

\usepackage[ruled,vlined,linesnumbered]{algorithm2e}
\usepackage{amssymb}
\usepackage{bm}
\usepackage{bbding}
\usepackage{booktabs} 
\usepackage{tabularx}
\usepackage{multirow}
\usepackage{etoolbox,siunitx}
\usepackage{color}      
\usepackage{diagbox}
\usepackage[font=scriptsize,skip=2pt,subrefformat=parens]{subcaption}

\usepackage{xspace}

\usepackage[colorlinks,linkcolor=red,anchorcolor=blue,citecolor=blue,CJKbookmarks=True]{hyperref}
\usepackage{natbib}

\def\name{SLRNet}

\newcommand{\Eq}{Eq.\@\xspace}

\newcommand{\Fig}{Fig.\@\xspace}

\newcommand{\Tab}{Tab.\@\xspace}

\newcommand{\Sec}{Sec.\@\xspace}
\newcommand{\Alg}{Algorithm\@\xspace}

\newcommand{\norm}[1]{\left\lVert#1\right\rVert}

\newcommand{\prelim}[1]{{#1}}

\newlength\savewidth
\begin{document}

\title{Learning Self-Supervised Low-Rank Network for Single-Stage Weakly and Semi-Supervised Semantic Segmentation}

\author{
        Junwen Pan$^*$         \and
        Pengfei Zhu$^{*\dagger}$        \and
        Kaihua Zhang       \and
        Bing Cao            \and
        Yu Wang            \and
        Dingwen Zhang      \and
        Junwei Han      \and
        Qinghua Hu
}

\institute{
            Junwen Pan \at
              Tianjin University, Tianjin, China \\
              \email{junwenpan@tju.edu.cn}           
           \and
            Pengfei Zhu \at
              Tianjin University, Tianjin, China \\
              \email{zhupengfei@tju.edu.cn} \\
              $\dagger$ Corresponding author \\
              $*$ These authors contributed equally.
          \and
          Kaihua Zhang \at
             Nanjing University of Information Science and Technology, Nanjing, China \\
             \email{zhkhua@gmail.com} 
          \and
          Bing Cao, Yu Wang and Qinghua Hu \at
             Tianjin University, Tianjin, China 
          \and
          Dingwen Zhang and Junwei Han \at
            Northwestern Polytechnical University, Xi’an, China 
}

\date{Received: date / Accepted: date}

\maketitle
\begin{abstract}
  Semantic segmentation with limited annotations, such as weakly supervised semantic segmentation (WSSS) and semi-supervised semantic segmentation (SSSS), is a challenging task that has attracted much  attention recently.
  Most leading WSSS methods employ a sophisticated multi-stage training strategy to estimate pseudo-labels as precise as possible, but they suffer from high model complexity.
  In contrast, there exists another research line that trains a single network with image-level labels in one training cycle.
  However, such a single-stage strategy often performs poorly because of the compounding effect caused by inaccurate pseudo-label estimation.
  To address this issue, this paper presents a Self-supervised Low-Rank Network (SLRNet) for single-stage WSSS and SSSS.
  The SLRNet uses cross-view self-supervision, that is, it simultaneously predicts several complementary attentive LR representations from different views of an image to learn precise pseudo-labels.
  Specifically, we reformulate the LR representation learning as a collective matrix factorization problem and optimize it jointly with the network learning in an end-to-end manner.
  The resulting LR representation deprecates noisy information while capturing stable semantics across different views, making it robust to the input variations, thereby reducing overfitting to self-supervision errors.
  The SLRNet can provide a unified single-stage framework for various label-efficient semantic segmentation settings: 1) WSSS with image-level labeled data, 2) SSSS with a few pixel-level labeled data, and 3) SSSS with a few pixel-level labeled data and many image-level labeled data.
  Extensive experiments on the Pascal VOC 2012, COCO, and L2ID datasets demonstrate that our SLRNet outperforms both state-of-the-art WSSS and SSSS methods with a variety of different settings, proving its good generalizability and efficacy.
\end{abstract}

\section{Introduction}
\label{intro}

\begin{figure}[!t]
   \centering
  \def\svgwidth{\linewidth}
    \includegraphics[width=1.0\linewidth]{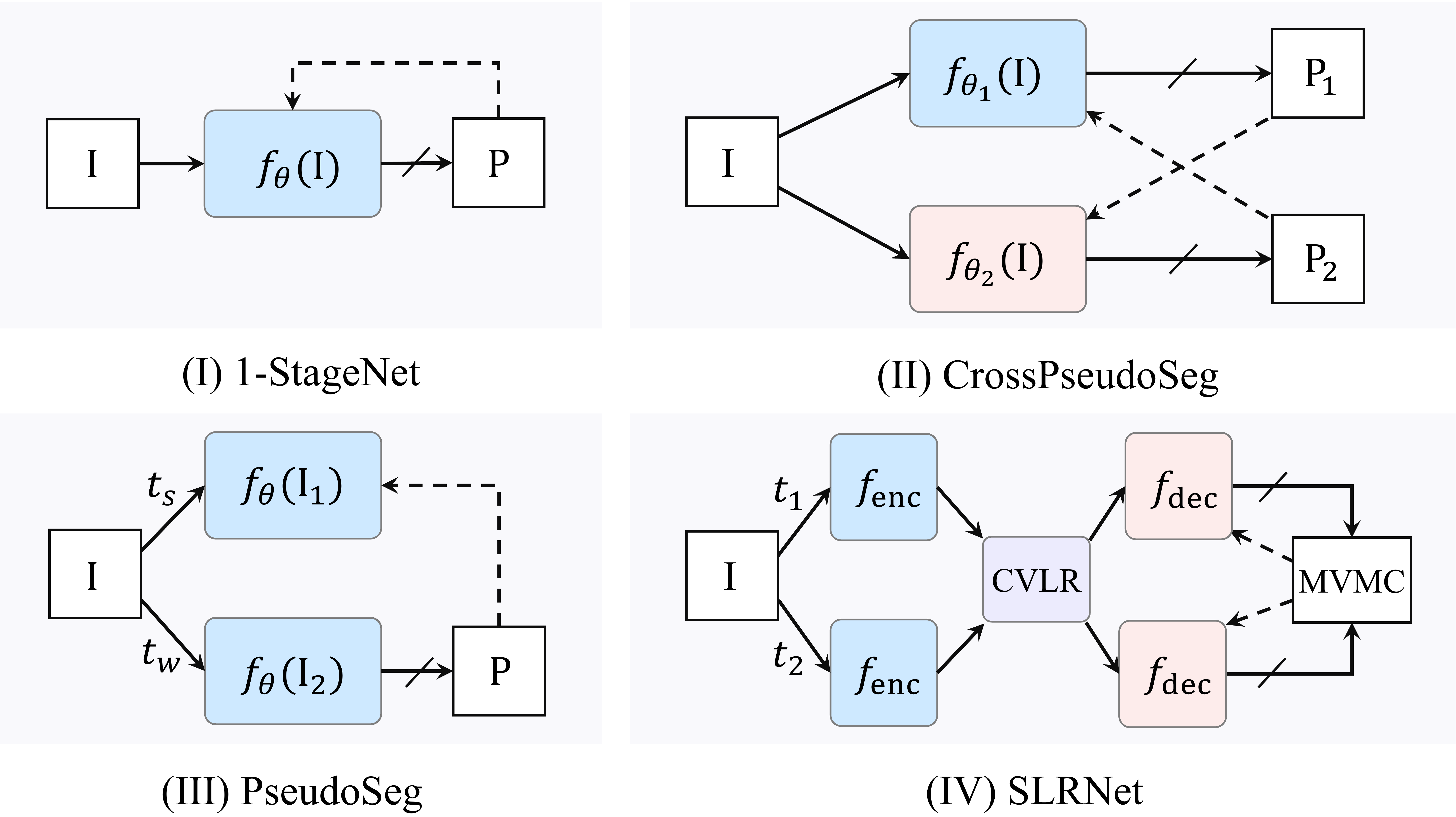}
  \vspace{-0.5cm}
  \caption{
    Overview of pseudo supervision architectures: (I) Single pseudo supervision for WSSS ~\citep{Araslanov020:SingleStage}, (II) Cross pseudo supervision for SSSS ~\citep{Chen2021:CrossPseudo}, (III) PseudoSeg supervision for SSSS ~\citep{zou2020:pseudoseg}, and (IV) The SLRNet with MVMC and CVLR can mitigate the compounding effect of pseudo supervision error. `$\rightarrow$' means the forward operation, `$\dashrightarrow$' means pseudo supervision,  `/' on `$\rightarrow$' means stop-gradient and pseudo-label generation. $t$ denotes the transformation operator, $s$ and $w$ are short for ``strong'' and ``weak'' respectively. 
  }
   \vspace{-1em}
  \label{fig:overview}
\end{figure}

Semantic segmentation is a fundamental computer vision task that aims to assign a label to each pixel, promoting the development of many downstream tasks, such as scene parsing, autonomous driving, and medical image analysis~\citep{deeplabv3plus2018,ZhouZPXFBT19:ADE20K,HavaeiDWBCBPJL17:BrainTumor}.
Recently, deep learning based semantic segmentation models~\citep{LongSD15:FCN,deeplabv3plus2018}, trained with large-scale data labeled at pixel level, have achieved impressive progress.
However, such supervised approaches require intensive manual annotations that are time-consuming and expensive, which have inspired many investigations about learning with low-cost annotations,
such as semi-supervised semantic segmentation (SSSS) with limited amounts of labeled data, weakly supervised semantic segmentation (WSSS) with bounding boxes~\citep{dai2015:boxsup}, scribbles~\citep{lin2016:scribblesup}, points~\citep{bearman2016s:pointsup}, and image-level labels~\citep{KolesnikovL16:SEC}.
Nevertheless, there is a considerable gulf between weakly supervised and semi-supervised approaches.

Most popular image-level WSSS methods~\citep{AhnCK19:IRN,dong_2020:conta,SunWDG20:MCIS} resort to multiple training and refinement stages to obtain more accurate pseudo-labels while avoiding error accumulation.
These methods often start from a weakly supervised localization, such as a class activation map (CAM)~\citep{ZhouKLOT16:CAM}, which highlights the most discriminative regions in an image.
In this approach, diverse enhanced CAM-generating networks~\citep{LeeKLLY19:FickleNet,WangZKSC20:SEAM,SunWDG20:MCIS} and CAM-refinement procedures ~\citep{AhnK18:PSA,AhnCK19:IRN,Shimoda2019:SSDD} have been designed to expand the highlighted area to the entire object or eliminate the wrongly highlighted area.
Although these multi-stage methods can produce more accurate pseudo-labels, they suffer from the need of a large number of hyper-parameters and complex training procedures.
Single-stage WSSS methods ~\citep{Zheng15:CRFRNN,PapandreouCMY15:EM} have received less attention because their segmentation is less accurate than that of multi-stage methods.
Recently, ~\cite{Araslanov020:SingleStage} proposed a simple single-stage WSSS model that generates pixel-level pseudo-labels online as self-supervision (\Fig\ref{fig:overview} (I)). However, its accuracy is still not comparable with that of multi-stage approaches.
In contrast, the simple online pseudo-supervision scheme has made promising progress in SSSS (\Fig\ref{fig:overview}(II)~\citep{Chen2021:CrossPseudo} and (III)~\citep{zou2020:pseudoseg}).

We argue that the cause of the inferior performance of the online pseudo supervised WSSS is the compounding effect of errors caused by online inaccurate pseudo supervision.
Like multi-stage refinements, online pseudo-label supervision should gradually improve the semantic fidelity and completeness during the training process.
However, this also increases the risk that errors are mimicked and accumulated with the gradient flows being backpropagated from the top to the lower layers.
Consistency learning is widely used as additional supervision to semi-supervised learning~\citep{OualiHT20:CCT,Chen2021:CrossPseudo}. 
However, in practice, existing consistency-based methods are not applicable to image-level weakly supervised settings. 
First, they require pixel-level supervision to avoid the collapsing solution~\citep{Xinlei:SimSiam}. Second, the dominance of consistency harms the region expansion for WSSS.

To this end, we propose the \textbf{S}elf-supervised \textbf{L}ow-\textbf{R}ank \textbf{Net}work (SLRNet) for single-stage WSSS and SSSS.
\prelim{
As illustrated in \Fig\ref{fig:overview}(IV),
the SLRNet simultaneously predicts several segmentation masks  for various augmented versions of one image, which are jointly calibrated and refined by a multi-view mask calibration (MVMC) module to generate one pseudo-mask for self-supervision.
The pseudo-mask leverages the complementary information from various augmented views, which enforces the cross-view consistency on the predictions.
To further regularize the network, the SLRNet introduces the LR inductive bias implemented by a cross-view low-rank (CVLR) module. 
The CVLR exploits the collective matrix factorization to jointly decompose the learned representations from different views into sub-matrices while recovering a clean LR signal subspace.
Through the dictionary shared over different views, a variety of related features from various views can be refined and amplified to eliminate the ambiguities or false predictions.
Thereby, the input features of the decoder deprecate noisy information, and this can effectively prevent the network from overfitting to the false pseudo-labels.
Additionally, instead of directly randomly initializing the sub-matrices, a latent space regularization is designed to improve the optimization efficiency.
}

\prelim{
The SLRNet is an efficient and elegant framework that generalizes well to different label-efficient segmentation settings without additional training phases.
For instance, to simultaneously utilize image-level and pixel-level labels, previous SSSS methods~\citep{LeeKLLY19:FickleNet,WeiXSJFH18:mdc} have to generate and refine pseudo-labels \textit{offline} using WSSS model, which are bundled with pixel-level labels to train a network in the next stage.
Such a multi-stage scheme provides a marginal improvement over dedicated SSSS algorithms~\citep{OualiHT20:CCT,zou2020:pseudoseg} with unlabeled data.
In contrast, the SLRNet directly introduces additional pixel-level supervision while combining it with image-level data without extra cost.
In other words, the online pseudo-mask generation takes into account both image-level and pixel-level labels in a single training phase and is undoubtedly more accurate.
To the best of our knowledge, the SLRNet is the first attempt to bridge these tasks into a unified single-stage scheme, allowing it to maximize exploiting various annotations with a limited budget. 
}

In our experiments, we first validate the performance of SLRNet in an image-level WSSS setting on several datasets, including Pascal VOC 2012~\citep{EveringhamGWWZ10:VOC}, COCO~\citep{LinMBHPRDZ14:COCO}, and L2ID~\citep{YunchaoWei2020:lid20}.
Extensive experiments demonstrate that the cross-view supervision and the CVLR help improve semantic fidelity and completeness of the generated segmentation masks.
Notably, the SLRNet also establishes new state-of-the-arts for various label-efficient semantic segmentation tasks, including 1) WSSS with image-level labeled data, 2) SSSS with pixel-level and image-level labeled data and 3) SSSS with pixel-level labeled and unlabeled data.
Moreover, the SLRNet achieves the best performance at the WSSS Track of CVPR 2021
Learning from Limited and Imperfect Data (L2ID) Challenge~\citep{YunchaoWei2020:lid20}, outperforming other competitors by large margins of $\sim 9.35\%$ in terms of mIoU.

The main contributions of this work are summarized as follows:
\begin{itemize}
  \item[1)] We propose an effective cross-view self-supervision scheme, incorporating the CVLR module, to alleviate the compounding effect of self-supervision errors for the online pseudo-label training.
  \item[2)] We present a plug-and-play collective matrix factorization method with latent space regularization for multi-view LR representation learning, which can be readily embedded into any Siamese networks for end-to-end training.   \item[3)] The SLRNet provides a unified framework that can be well generalized to learn a segmentation model from different limited annotations in various WSSS and SSSS settings.
  \item[4)] The SLRNet achieves leading performance compared to a variety of state-of-the-art methods on Pascal VOC 2012, COCO, and L2ID datasets for both WSSS and SSSS tasks.
\end{itemize}

\section{Related Work}
This section reviews a variety of methods related to the proposed {\name}, including the WSSS, the SSSS, the LR representation, and the self-supervised learning methods.
\subsection{Weakly Supervised Semantic Segmentation}
\label{sec:wsss_related}
In the past years, various variants of WSSS methods have been developed and evolved rapidly, which can be categorized into multi-stage and single-stage classes.
Most multi-stage WSSS methods with image-level supervision start from the CAM~\citep{ZhouKLOT16:CAM}.
These methods refine the CAM obtained from a pre-trained CAM-generating (classification) network to generate segmentation pseudo-labels, with the aim of expanding the highlighted area to the entire object or eliminating the false highlighted area.
The prevailing method is to consider the semantic completeness and fidelity of the seed region when training the CAM-generating network using, for instance, atrous convolution ~\citep{WeiXSJFH18:mdc}, stochastic feature selection~\citep{LeeKLLY19:FickleNet}, the idea of erasing ~\citep{WeiFLCZY17:AdverErasing,HouJWC18:SelfErasing}, cross-image affinity ~\citep{SunWDG20:MCIS}, and the equivariant for various scaled input images~\citep{WangZKSC20:SEAM}.
Although the seed obtained from an improved CAM-generating network is better, most of these methods still need extra CAM-refinement procedures, such as random walk~\citep{AhnK18:PSA}, region growing ~\citep{HuangWWLW18:DSRG}, or an additional network for distillation.
Moreover, some of these methods utilize class-agnostic saliency maps to obtain background cues.
The multi-stage methods use a series of algorithms to improve the WSSS accuracy by carefully tuning the hyperparameters of each stage, leading to a rapid increase in complexity.

In contrast to the multi-stage methods, the single-stage approaches train WSSS models using only one training cycle ~\citep{PinheiroC15:fromimage,PapandreouCMY15:EM,Zheng15:CRFRNN}, but they cannot perform favorably because of their inferior segmentation accuracy.
Recently, ~\citep{Araslanov020:SingleStage} proposed a simple yet effective single-stage model, \textit{i.e.}, they train a segmentation network with image-level labels and produce refined masks online as self-supervision. However, this self-trained model is still unable to compete with the latest multi-stage methods~\cite{AhnCK19:IRN,LeeKLLY19:FickleNet,WangZKSC20:SEAM} in accuracy.

\subsection{Semi-Supervised Semantic Segmentation}

Generally, in semi-supervised learning, only a small subset of training images are assumed to have annotations, and a large number of unlabeled data are exploited to improve performance.
Early SSSS models~\citep{HungTLL018:AdvSemSeg,SoulySS17:GANSeg} based on generative adversarial networks learn a discriminator between the prediction and the ground truth mask  or generate additional training data.
Recently, consistency based approaches have been extensively explored. These approaches enforce the predictions to be consistent, either using transformed input images~\citep{FrenchLAMF20:SSSSNeedsPerturbations,zou2020:pseudoseg}, perturbed feature representations~\citep{OualiHT20:CCT}, or different networks~\citep{Chen2021:CrossPseudo}.
PseudoSeg~\citep{zou2020:pseudoseg} and CrossPseudo~\citep{Chen2021:CrossPseudo} realize the idea of consistency by designing pseudo-labels online to encourage the cross-view consistency, \textit{i.e.}, one view generates pseudo-labels for supervising another view, instead of explicitly enforcing the prediction consistency.
In this paper, our experiments illustrate that this explicit consistency impairs the expansion of highlighted regions under a weakly supervised setting.

Another WSSS-based line of research involves harnessing low-cost image-level supervision. As described in \Sec\ref{sec:wsss_related}, WSSS approaches generate segmentation pseudo-labels that can then be used to train a segmentation network together with the human-annotated pixel-level labels~\citep{LeeKLLY19:FickleNet,WeiXSJFH18:mdc,LiWPE018:GAIN}.
In contrast, our SLRNet provides a unified single-stage WSSS and SSSS framework without offline pseudo-label generation or multi-stage training.

\subsection{Low-Rank Representation}
The LR representation seeks a compact data structure and has been widely applied to subspace clustering~\citep{liu2012robust}, dictionary learning~\citep{ma2012sparse},
matrix decomposition~\citep{cabral2013unifying},
and deep network approximation~\citep{tai2016convolutional}.
Liu et al. proposed a robust subspace clustering model using an LR representation~\citep{liu2012robust};
Ma et al. proposed a discriminative LR dictionary learning algorithm for face recognition~\citep{ma2012sparse}; and Ricardo et al. proposed a unified approach to bilinear factorization and nuclear norm regularization for LR matrix decomposition~\citep{cabral2013unifying}.
Because of the redundancy of convolutional filters, LR regularization is imposed to speed up
convolutional neural networks~\citep{tai2016convolutional}.
\prelim{
Recently, \cite{Zheng:ham} and ~\cite{Lixia19:EMANET} introduced the LR reconstruction into segmentation as an alternative to the self-attention mechanism.
Both of them merely endeavor to denoise variance and capture the invariant representations between related pixels in a single feature map, while ignoring the inconsistency and noise in cross-view feature maps, and thus can bring a rather marginal effect towards alleviating the error accumulation in those settings with limited supervision.
}
\subsection{Self-Supervised Learning}
\prelim{
Self-supervised learning aims at designing pretext tasks to learn general feature representations from large-scale unlabeled data without human-annotated labels.
Classical pretext tasks include image and video generation~\citep{PathakKDDE16:context,LedigTHCCAATTWS17:PhotoRealistic}, spatial or temporal context learning~\citep{DoerschGE15:contextPred,LeeHS017:sortingSeq}, and free semantic label-based methods~\citep{StretcuL15:mfmodv}.
Recently, contrastive learning has become popular, whose core idea is to attract positive pairs and repulse negative pairs~\citep{ChenK0H20:SimCLR,He0WXG20:moco}.
Siamese architectures for contrastive learning can model transformation invariance by weight-sharing, \textit{i.e.}, two views of the same sample should produce the consistent outputs ~\citep{Xinlei:SimSiam}.
The pre-trained features by self-supervised methods are transferred to the downstream tasks for further learning.
For downstream tasks requiring dense predictions (\textit{e.g.}, segmentation and detection), there are also a flurry of recent work~\citep{Xie_2021_ICCV:DetCo,XieL00L021:PixPro,WangZSKL21:DenseCL,PinheiroABGC20:VADeR} exploring pixel-level contrastive learning that model local invariance by considering the correspondence between local representations.
}

\prelim{
Considering the label-efficient segmentation with limited supervision signal, it is intuitive to introduce self-supervision as additional constraint to narrow the gap between label-efficient and fully-supervised settings.
SEAM~\citep{WangZKSC20:SEAM} exploited a Siamese network to model scale equivariance, while CIAN~\citep{FanZTSX20:cian} and MCIS~\citep{SunWDG20:MCIS} mined the cross-image affinity from image pairs.
Here, self-supervision is manifested in distinct dimensions: along with pixel-level prediction consistency over different transformed views, free pseudo ground-truth calibrated by the MVMC can also be regarded as self-supervision.
Additionally, the CVLR enforces consistency between cluster assignments produced for different views, which is also related to self-supervised SWaV~\citep{CaronMMGBJ20:SWaV}.
}

\begin{figure*}[t]
    \def\svgwidth{\linewidth}
    \input{figures/framework/overall2.tex}
    \caption{\prelim{
        \textbf{Left:} Architecture of the proposed SLRNet.
        The SLRNet is a multi-branch network with shared parameters, which simultaneously predicts the masks of multiple views ($v$ and $u$) from one image.
        The MVMC generates a pseudo segmentation mask as self-supervision (using loss $\mathcal{L}_{seg}$) by calibrating the false activations in these predicted masks.
        Meanwhile, the outputs are also constrained by image-level loss $\mathcal{L}_{cls}$ and explicit cross-view regularization $\mathcal{L}_{reg}$.
        \textbf{Right:} The CVLR module.
        The CVLR is implemented by the cross-view matrix factorization, which reduces the multi-view features into a shared dictionary ($\mathbf{D}$) and a set of code matrices ($\mathbf{C}^{v}$), where the noises are removed during the reconstruction.
        }
    }
    \label{fig:framework}
\end{figure*}
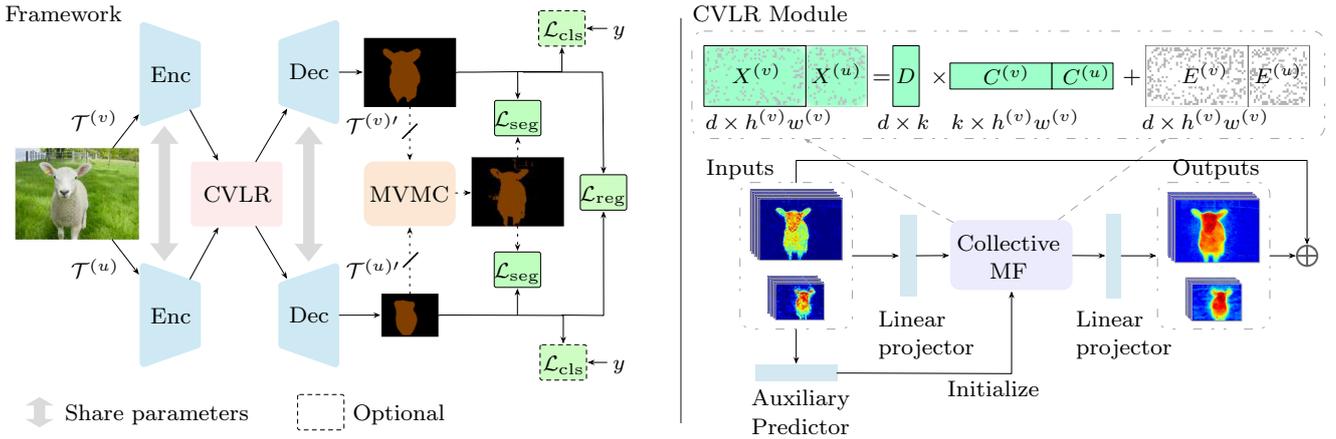

\section{Methodology}
\label{sec:approach}

This section introduces the proposed SLRNet in details. First, we introduce the unified framework of SLRNet for label-efficient semantic segmentation in \Sec\ref{sec:framework}. Then, we introduce how to design the cross-view supervision in \Sec\ref{sec:cv_sup}. Among it, to reduce the compounding effect caused by self-supervision errors, the MVMC method is proposed to provide cross-view pseudo-label supervision.
Finally, \Sec\ref{sec:cross_view_lowrank} introduces the CVLR model, among which we introduce the inductive bias of the LR property into the neural network using the collective matrix factorization method to further mitigate the compounding effect of errors. 
The CVLR module is then integrated into the network for end-to-end training.

\subsection{Framework of the SLRNet}
\label{sec:framework}
Similar to typical fully-supervised semantic segmentation, the SLRNet consists of one network trained in one stage without complicated steps for both WSSS and SSSS tasks in a unified framework.
Specifically, the SLRNet expands the established encoder--decoder segmentation network~\citep{deeplabv3plus2018} into a simple shared-weight Siamese structure (\Fig\ref{fig:framework} left).
The SLRNet takes $m$ views $\mathbf{I}^{(v)}$ from an image $\mathbf{I}$ augmented by transformations $\mathcal{T}^{(v)}$ as input.
For explanation clarity, the superscript $v$ denotes the index of view $v\in \mathcal{V}$ and $|\mathcal{V}|$ denotes the total number of views.
The encoder network processes these views and produces feature maps $\mathbf{X}^{(v)} \in \mathbb{R}^{h^{(v)} \times w^{(v)} \times d}$.
The CVLR module  in \Sec\ref{sec:cross_view_lowrank} jointly factorizes the high-dimensional noisy features $\mathbf{X}^{(v)}$ from different views into sub-matrices and reconstructs the LR features $\tilde{\mathbf{X}}^{(v)}$.
Afterwards, the features with the LR property are fed to the decoder to predict the segmentation logits $\hat{\mathbf{M}}^{(v)}$.

From a unified perspective, we consider three typical types of data under established label-efficient settings: pixel-level labeled data $\mathcal{D}_p\in \{ (\mathbf{I}_i, \mathbf{M}_i) \}_{i=1}^{|\mathcal{D}_c|}$, image-level (\textit{i.e.} classification) labeled data $\mathcal{D}_c\in \{ (\mathbf{I}_i, \mathbf{y}_i) \}_{i=1}^{|\mathcal{D}_c|}$, and unlabeled data $\mathcal{D}_u\in \{\mathbf{I}_i\}_{i=1}^{|\mathcal{D}_u|}$.
During training, the samples in $\mathcal{D}_p$ use the manually labeled pixel-wise mask $\mathbf{M}$, and those in $\mathcal{D}_c$ and $\mathcal{D}_u$ use the estimated pseudo-labels $\tilde{\mathbf{M}}$ produced by the MVMC module described in \Sec\ref{sec:cv_sup} as ground-truth to construct the following pixel-level cross entropy (CE) loss:
\begin{equation}
  \begin{aligned}
  \mathcal{L}_\text{seg} & =
    \sum_{i \in \mathcal{D}_p} \sum_{v\in \mathcal{V}} \mathcal{L}_\text{CE}(\hat{\mathbf{M}}_{i}^{(v)},{\mathbf{M}}_{i}) \\
    &+ \sum_{i \in \mathcal{D}_c \cup \mathcal{D}_u} \sum_{v\in \mathcal{V}}\mathcal{L}_\text{CE}(\hat{\mathbf{M}}_{i}^{(v)},{\tilde{\mathbf{M}}}_{i}).
  \end{aligned}
  \label{eq:seg_loss}
\end{equation}
We apply normalized global weighted pooling with focal mask penalty ~\citep{Araslanov020:SingleStage} on the mask logits $\mathbf{M}^{(v)}$ to obtain class scores $\mathbf{\hat{y}}^{(v)}=pool(\mathbf{M}^{(v)})$.
Besides, we employ the binary cross entropy (BCE)~\citep{paszke2017:pytorch} for multi-label one-versus-all classification defined as follows:
\begin{equation}
    \mathcal{L}_\text{cls}= \sum_{i \in \mathcal{D}_c} \sum_{v\in\mathcal{V}} \mathcal{L}_\text{BCE}\left(\mathbf{\hat{y}}_{i}^{(v)},\mathbf{y}_{i}\right) .
    \label{eq:cls_loss}
\end{equation}

\subsection{Cross-view Supervision}
\label{sec:cv_sup}
The SLRNet employs pixel-level pseudo-labels generated online for self-supervision.
How to generate the desired pseudo-mask $\tilde{\mathbf{M}}$ is a pivotal question.
A naive solution is to simply utilize the decoder output of a trained model after confidence thresholding, as suggested by ~\cite{ZophGLCLC020:RethinkingPreTraining,SohnBCZZRCKL20:FixMatch}.
However, in a label-efficient segmentation regime, especially in image-level WSSS, the generated hard/soft pseudo-mask is fairly coarse.
As the network becomes deeper and deeper, errors are prone to be accumulated as the gradient flows are backpropagated from the top to the lower layers, and thus yield inferior performance ~\citep{deeplabv3plus2018,Araslanov020:SingleStage}.
We use two distinct yet efficient insights to generate better pseudo-label masks: First, we use the complementary information through multi-view fusion to eliminate potential errors in the decoder outputs; Second, we utilize  explicit and implicit cross-view supervision to regularize the network to produce more consistent outputs.

\paragraph{Multi-view Mask Calibration.}
Most previous WSSS approaches employ \textit{offline} multi-scale ensemble (\textit{i.e.} test-time augmentation) and complicated post-processing steps (\textit{e.g.} \cite{AhnK18:PSA,AhnCK19:IRN,dong_2020:conta}) to refine the coarse outputs as pseudo-labels. 
These approaches require multi-stage training that increases model complexity.
Here, we present an efficient and effective \textit{online} MVMC scheme for mask calibration.
In the early training steps, the network's output is prone to activate partial regions of interest or too many background regions, whereas we can use the output of different augmented versions to calibrate the false predictions.
Specifically, the network simultaneously predicts a set of masks $\{\mathbf{M}^{(v)} \in \mathbb{R}^{h^{(v)} \times h^{(v)}  \times d} \}_{v\in\mathcal{V}}$ for various transformed versions of the same image whose spatial pixels are not aligned because of various geometric transformations.
First, the output $\hat{\mathbf{M}}^{(v)}$ is transformed respectively by the inverse geometric transformations $\mathcal{T}^{(v)\prime}$. Note that we assume that $\mathcal{T}^{(v)\prime}$ is ``differentiable'', \textit{e.g.}, we use bilinear interpolation and flipping. Then, the fused masks are processed by the refinement procedure $\mathcal{R}$.
The whole calibration process can be formulated as:
\begin{equation}
  \tilde{\mathbf{M}} = \mathcal{R}\left(softmax \left(\frac{\sum_{v\in\mathcal{V}} \mathcal{T}^{(v)\prime}(\mathbf{M}^{(v)}) } {|\mathcal{V}|} \right)\right).
\end{equation}
Because a classic refinement algorithm such as dense CRF~\citep{KrahenbuhlK11:CRF} slows down the training process, we refine the coarse masks with respect to appearance affinities through pixel-adaptive convolution~\citep{su2019:PAC,Araslanov020:SingleStage}.
To generate the one-hot hard pseudo-labels, we retain the pseudo-labels of pixels with confidence higher than a threshold $\gamma$, and ignore the pixels with low confidence or those belonging to multiple categories.

\paragraph{Implicit and Explicit Cross-view Supervision.}
The pseudo-masks produced by the MVMC utilize the complementary information from different views, and hence the pseudo supervised segmentation loss in \Eq\ref{eq:seg_loss} implicitly enforces prediction consistency.
In contrast to PseudoSeg~\citep{zou2020:pseudoseg} and CPS~\citep{Chen2021:CrossPseudo}, which explicitly realize cross-view supervision, the proposed approach implicitly implements cross-view consistency regularization and exploits the MVMC to filter out false pseudo-labels.
Extensively investigations reveal that pure cross-view supervision is prone to cause mode collapse in unsupervised settings~\citep{Xinlei:SimSiam,ChenK0H20:SimCLR}.
In practice, our experiments in \Sec\ref{sec:ablation} also show that the dominance of explicit cross-view supervision can compromise the semantic completeness in WSSS.
Therefore, a proper and adjustable cross-view supervision strength is crucial in the settings with limited supervision.
Here, we define an explicit consistency loss, which is tuned by a scaling factor $\lambda_\text{reg}$, to compensate for the implicit cross-view consistency:
\begin{equation}
\label{eq:reg_mask}
\begin{aligned}
    & \mathcal{L}_\text{reg}^\text{mask} = \\
    & \sum_{i \in \mathcal{D}}
    \sum_{v,u\in \mathcal{V}, v\neq u}
    \sum_{c\in \mathbf{y}_i}
    s{\left[\mathcal{T}^{(v)\prime}\left(\hat{\mathbf{M}}^{(v)}_{i,c}\right), \mathcal{T}^{(u)\prime}\left(\hat{\mathbf{M}^{(u)}_{i,c}}\right)\right]},
\end{aligned}
\end{equation}
where $s(\cdot,\cdot)$ is a distance measure between two output masks, $\mathcal{D} = \mathcal{D}_p \cup  \mathcal{D}_c \cup  \mathcal{D}_u$ denotes all samples, and $(v,u)$ denotes the pairs composed of different views. We consider only the mask of category $c$ contained in the classification label $\mathbf{y}_i$ for the $i$-th example. 
For unlabeled sample $i\in \mathcal{D}_u$, all categories are involved in the computation.
\subsection{Cross-view Low-Rank Module}
\label{sec:cross_view_lowrank}

\prelim{Although the MVMC can provide more accurate pseudo-label mask and regularization effect, there still exist a lot of clutter and incompleteness in pseudo-masks. 
To further reduce the compounding effect of self-supervision errors, we introduce an additional inductive bias, \textit{i.e.}, the LR property of the cross-view deep embeddings, and formulate it into an optimization objective in terms of collective matrix factorization.
\Fig\ref{fig:framework} (right) illustrates the architecture of the CVLR module.
Its essence lies in capturing the invariant LR feature representations from differently transformed views to reduce the accumulated errors introduced by distorted and noisy self-supervision.
}

\paragraph{Matrix Factorization (MF).}
Before introducing the proposed CVLR module, we first review the key concept of LR MF.
The MF reduces a matrix into constituent parts, which disentangles latent structures in high-dimensional complex data.
Given $n$ data features of dimension $d$ denoted as $\mathbf{X}=[\mathbf{x}_1;\ldots;\mathbf{x}_n] \in \mathbb{R}^{d \times n} $, we assume that there is a low-dimensional subspace hidden in $\mathbf{X}$. Therefore, $\mathbf{X}$ can be decomposed into a dictionary matrix  $\mathbf{D}=[\mathbf{d}_1;\ldots;\mathbf{d}_k] \in \mathbb{R}^{d \times k}$ and an associated code matrix $\mathbf{C}=[\mathbf{c}_1;\ldots;\mathbf{c}_n] \in \mathbb{R}^{k \times n}$:
\begin{equation}
  \mathbf{X}=\tilde{\mathbf{X}}+\mathbf{E} = \mathbf{D}\mathbf{C}+\mathbf{E},
\end{equation}
where we denote the reconstructed feature as $\tilde{\mathbf{X}} \in \mathbb{R}^{n \times d} $ and the noise matrix as $\mathbf{E} \in \mathbb{R}^{n \times d}$. $\mathbf{E}$ is discarded in the reconstruction.
We assume $k \ll \min(n,d)$, thus $\tilde{\mathbf{X}}$ has the LR property:
\begin{equation}
rank(\tilde{\mathbf{X}}) \leqslant min(rank(\mathbf{D}), rank(\mathbf{C})) \leqslant k \ll min(d, n).
\end{equation}

\paragraph{Vector Quantization (VQ).}
VQ is a classic data compression method, which can be formulated as an optimization objective in terms of MF:
\begin{equation}
  \min_{D,C} \norm{ \mathbf{X} - \mathbf{D}\mathbf{C}}_{F} \quad s.t. \  {c}_{ij} \in \{0,1\},\  \sum_{i} {c}_{ij} = 1,
\end{equation}
where vector $\mathbf{c}_{j}$ is a one-hot encoding, which implies a crisp partitioning.

\begin{algorithm}[t!]
\SetAlgoLined
\KwIn{
  $ \mathbf{X} = \{\mathbf{X}^{(v)} \in \mathbb{R}^{d\times h^{(v)}w^{(v)}} \}_{v\in \mathcal{V}}$,
  $\mathbf{D} \in \mathbb{R}^{d\times k} $,
  $\mathbf{C} = \{\mathbf{C}^{(v)} \in \mathbb{R}^{k\times h^{(v)}w^{(v)}} \}_{v\in \mathcal{V}}$
}
\For{$t \leftarrow 1$ \KwTo $T$}{
  $\mathbf{D} \leftarrow \sum_{v\in \mathcal{V}} \mathbf{X}^{(v)} \mathbf{C}^{(v)\top} \mathbf{S}^{-1} $, 
  where $\mathbf{S} = diag(\sum_{v\in \mathcal{V}} \mathbf{C}^{(v)} \mathbf{1}_n)$\;
  $ \mathbf{C} \leftarrow \{ softmax \left( \frac{1}{\tau} \mathbf{D}^{\top}_\text{norm} \cdot \mathbf{X}^{(v)}, \text{axis}=0 \right) \}_{v\in \mathcal{V}}$,
  where $\mathbf{D}_\text{norm}=\left[ \frac{\mathbf{d}_1 }{\norm{\mathbf{d}_1}_2} ; ...; \frac{\mathbf{d}_k }{\norm{\mathbf{d}_k}_2}  \right]$ \;
}
 $\tilde{\mathbf{X}} \leftarrow \{ \tilde{\mathbf{X}}^{(v)} = \mathbf{D}\mathbf{C}^{(v)} \}_{v\in \mathcal{V}}$\;
\KwOut{ $\tilde{\mathbf{X}} $}
 \caption{Collective MF}
 \label{alg:cvlr}
\end{algorithm}

\paragraph{Collective MF.}
We investigate the idea of collective MF to learn shared latent factors from $|\mathcal{V}|$ matrices of multi-view features.
Different views use the same LR representation as part of the approximation, enabling feature sharing and interaction.
The objective function of Collective MF can be formalized as:
\begin{equation}
  \min_{D, \{ C^{(v)} \} } \sum_{v\in \mathcal{V}} \left(  \norm{ \mathbf{X}^{(v)} - \mathbf{D}\mathbf{C}^{(v)} }_{F} + r (\mathbf{C}^{(v)}) \right) ,
\end{equation}
where $r$ is the regularization term.
To minimize the objective function, we utilize the same alternating minimization method as VQ, which is also equivalent to K-means clustering~\citep{gray1998:quantization}.

We now describe a single iteration on a set of flattened deep feature matrices from various transformed views $\{ \mathbf{X}^{(v)} \in \mathbb{R}^{ d\times h^{(v)} \cdot w^{(v)} } \}_{v \in \mathcal{V}}$.
We exploit the weighted mean over varied augmented features to update the shared dictionary matrix $\mathbf{D}$ (Line 2 in \Alg\ref{alg:cvlr}), where the weight $\mathbf{C}^{(v)\top} \mathbf{S}^{-1}$ is calculated over the global codes.
Factor matrix $\mathbf{C}^{(v)} \in \mathbb{R}^{k \times h^{(v)}w^{(v)}}$ is computed via the softmax-normalized attention ~\citep{VaswaniSPUJGKP17:transformer} with temperature coefficient $\tau$ instead of the hard maximum, which enables the gradients to be backpropagated. 
Here we normalize each vector in $\mathbf{D}$ and set $\tau=1$ by default.
As described in \Alg\ref{alg:cvlr}, the CVLR updates the factor matrices $\mathbf{D}$ and $\{\mathbf{C}^{(v)}\}_{v\in \mathcal{V}}$ alternately. After $T$ iterations, the converged $\mathbf{D}$ and $\{\mathbf{C}^{(v)}\}_{v\in \mathcal{V}}$ are used to approximate input features $\{\mathbf{X}^{(v)}\}_{v\in \mathcal{V}}$.
As discussed above, the re-estimated matrices $\{ \tilde{\mathbf{X}}^{(v)} \}_{v\in \mathcal{V}}$ are endowed with the LR property.

\prelim{
With the above optimization, the features of various transformed views interact with each other iteratively and are compressed into a shared dictionary $\mathbf{D}$.
Then, these condensed semantics are propagated into individual views in the final reconstruction step.
From an intuitive perspective, as illustrated in \Fig\ref{fig:framework}, the re-estimated LR features amplify and refine the related features from complementary views, eliminate ambiguity or false responses, and thus produce more complete and clean activation regions.
}

\paragraph{Training with Back-Propagation.}
The above optimization is differentiable with respect to its parameters, which makes it suitable to incorporate with a convolutional neural network for end-to-end training.
However, the alternating minimization introduces recurrent neural network-like behavior, and randomly initialized factor matrices require multiple iterations, which will degrade performance due to the vanishing gradient in practice.
One possible solution is to stop the gradient flow in the iterations to avoid gradient instability and memorize the previous $\mathbf{D}$ to initialize the current optimization to reduce the number of iterations $T$~\citep{Lixia19:EMANET}.
In contrast, the gradient flow is allowed in CVLR and factor $\mathbf{C}^{(v)}$ is initialized with the feature map produced by the network.
Given feature map $\mathbf{Z}^{(v)} \in \mathbb{R}^{k\times h^{(v)} w^{(v)}}$,  $softmax(\mathbf{Z}^{(v)},axis=0)$ can be regarded as a good initialization of $\mathbf{C}^{(v)}$.
In contrast to random initialization, the optimization swiftly converges within a small number of iterations $T$.

\paragraph{Latent Space Regularization.}
For specific scenarios, there are diverse MF variants with various regularizations on $\mathbf{D}$ and $\mathbf{C}$, such as non-negativity~\citep{lee1999:NMF}, orthogonality~\citep{Ding:OrthogonalNMF} and latent space enforcement~\citep{HuTGL13:UnsupervisedSentiment}.
\prelim{
Here, although multi-view features are compressed into a shared dictionary, features with the same semantic meaning may still be assigned to different elements in $\mathbf{D}$. 
To this end, we regularize the latent space of code matrix $\mathbf{C}^{(v)}$ to be the pixel-level classification space by fully exploiting the limited supervision.
That is, we set the dimension $k$ of the factor matrices to the number of categories provide auxiliary supervision on the initialization of $\mathbf{C}^{(v)}$.
In this way, each element in $\mathbf{D}$ is assigned a specific meaning, ensuring cross-view invariance of the reconstructed LR representations.}
In addition, a cross-view regularization loss on the code matrix $\mathbf{C}$ is designed to encourage the cross-view consistency as follows:
\begin{equation}
\label{eq:reg_fact}
  \mathcal{L}_\text{reg}^\text{fact} = \sum_{v,u\in \mathcal{V}, i\neq j}
  s{\left(\mathcal{T}^{(v)\prime}\left(\mathbf{C}^{(v)}\right), \mathcal{T}^{(u)\prime} \left(\mathbf{C}^{(u)}\right)\right)},
\end{equation}
where $s(\cdot,\cdot)$ is a similarity measure, $\mathcal{T}^{(v)\prime}$ and $\mathcal{T}^{(u)^\prime}$ are inverse geometric transformations.

\paragraph{Detailed CVLR Module Design.}
\Fig\ref{fig:framework} presents the architecture of CVLR, which collaborates with the networks via two linear projectors and a skip connection.
Specifically, a learnable linear transformation maps the inputs to a feature space, and another one maps the approximation $\tilde{\mathbf{X}}$ to the input space.
To produce the feature $\mathbf{Z}^{(v)}$ as the initialization of $\mathbf{C}^{(v)}$, the auxiliary prediction head is composed of two convolutional layers supervised by the loss functions defined in \Eq\ref{eq:seg_loss} and \Eq\ref{eq:cls_loss}.

\section{Experiments}
\paragraph{Implementation Details.}
In our experiments, WideResNet-38~\citep{wu2019:WiderResNet} pre-trained on ImageNet~\citep{DengDSLL009:imagenet} and atrous spatial pyramid pooling (ASPP)~\citep{deeplabv3plus2018} form our encoder. The decoder consists of three convolutional layers and a stochastic gate~\citep{Araslanov020:SingleStage}, which mixes shallow and deep features.
We trained our model for 20 epochs with a stochastic gradient descent optimizer using a weight decay of $5\times10^{-4}$.
The learning rate was set to $5\times10^{-3}$ for randomly initialized parameters and $5\times10^{-4}$ for pre-trained parameters. 
We use L1 distance as the similarity measure $s(\cdot,\cdot)$ in \Eq\ref{eq:reg_mask} and \Eq\ref{eq:reg_fact}.
The final loss function can be expressed as $\mathcal{L}=\lambda_\text{seg} \mathcal{L}_\text{seg}+\lambda_\text{cls} \mathcal{L}_\text{cls}+\lambda_\text{reg} \mathcal{L}_\text{reg}^\text{mask}+\lambda_\text{reg} \mathcal{L}_\text{reg}^\text{fact}$.
In the first five epochs, the factors of the loss functions were
set to $\lambda_{seg}=0$ , $\lambda_{cls}=1$, and $\lambda_{reg}=4$. After that, they were set to the default values $\lambda_{seg}=1$ , $\lambda_{cls}=1$ and $\lambda_{reg}=4$.

\paragraph{Evaluation Metric.}
We report the results in terms of the mean of the class-wise intersection over union (mIoU) for all datasets.

\subsection{Experiment I: Learning WSSS from Pascal VOC Dataset}

\begin{figure*}[t]
    \begin{subfigure}[t]{0.5\linewidth}
        \centering
        \hspace{0.5cm}Ours\hspace{1.8cm} Ours+CRF \hspace{1cm}  Ground-truth
        \includegraphics[width=0.98\linewidth]{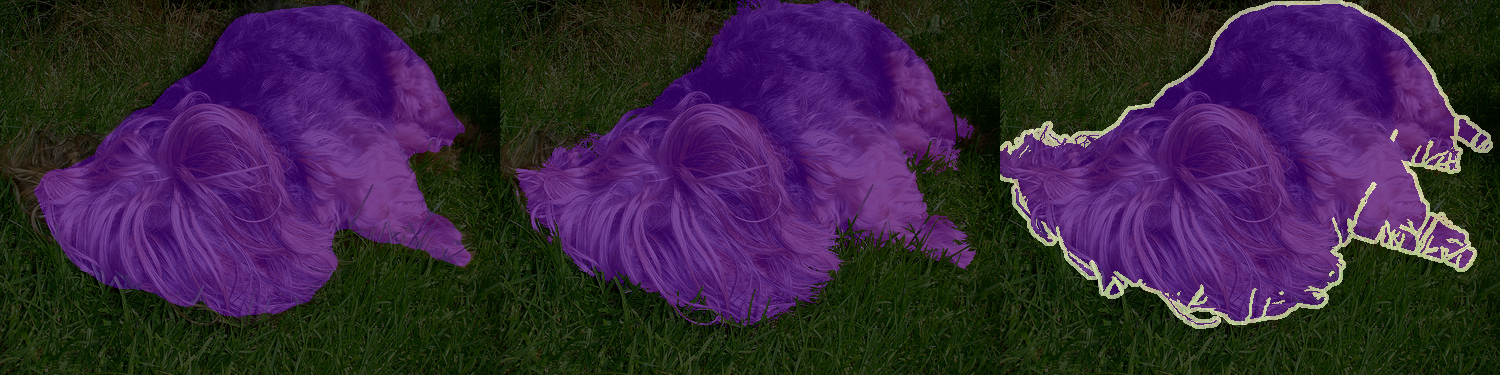}\\
        \includegraphics[width=0.98\linewidth]{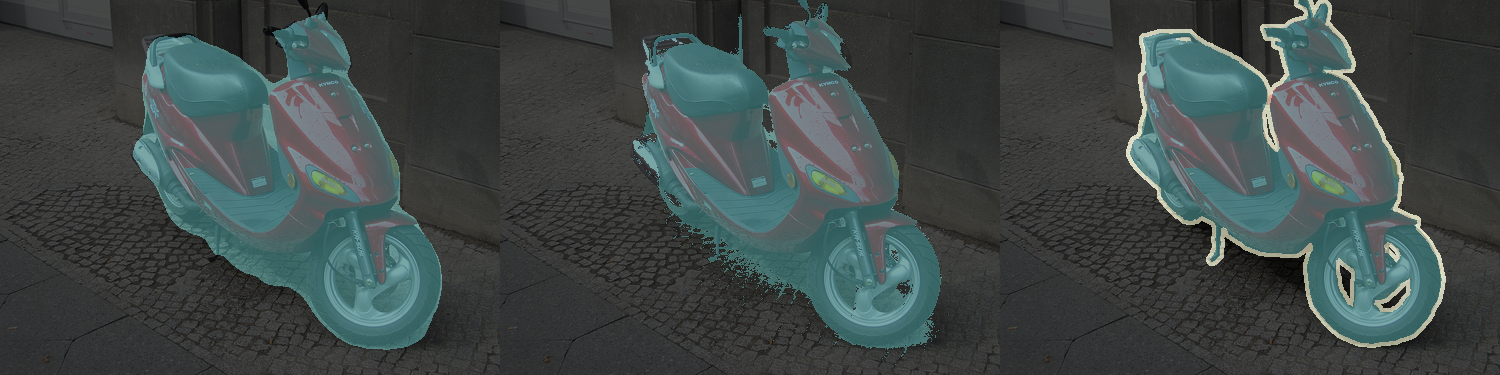}\\
        \includegraphics[width=0.98\linewidth]{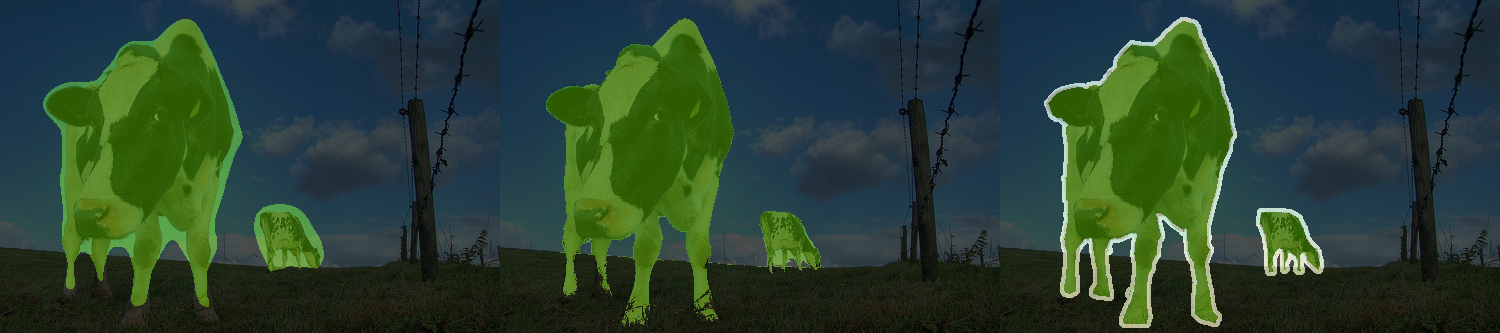}\\

        \includegraphics[width=0.98\linewidth,trim=0 0 0 10,clip]{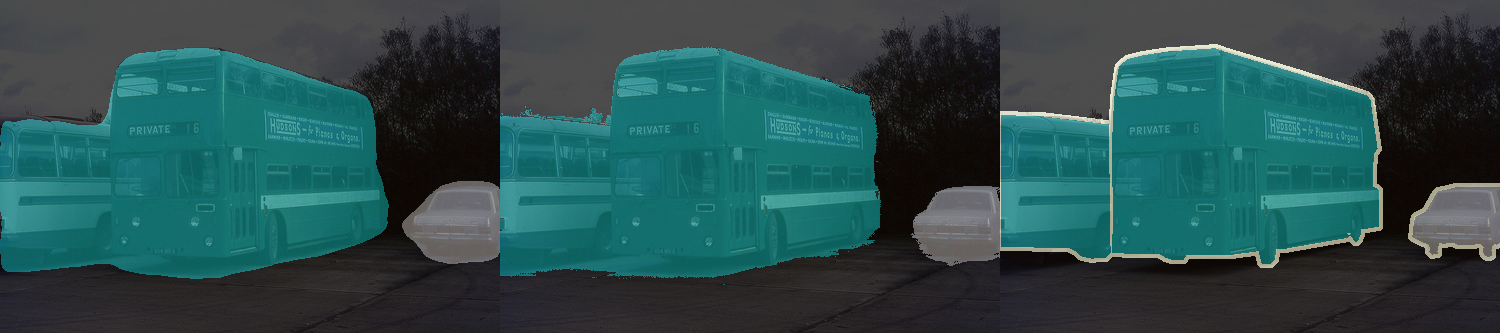}\\
        \includegraphics[width=0.98\linewidth,trim=0 0 0 25,clip]{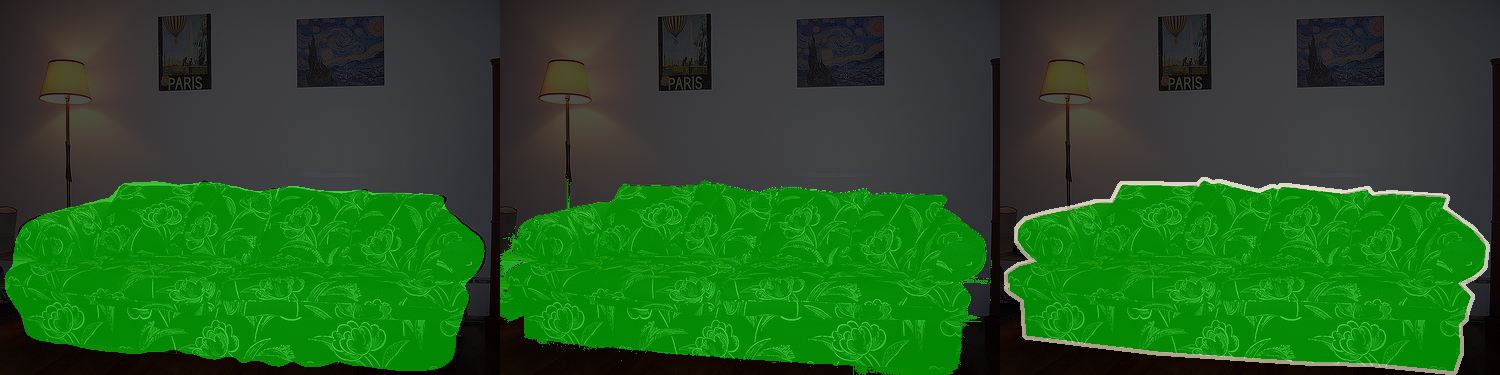}\\

    \end{subfigure}%
    \begin{subfigure}[t]{0.5\linewidth}
        \centering
        \hspace{0.5cm}Ours\hspace{1.8cm} Ours+CRF \hspace{1cm}  Ground-truth
        \includegraphics[width=0.98\linewidth]{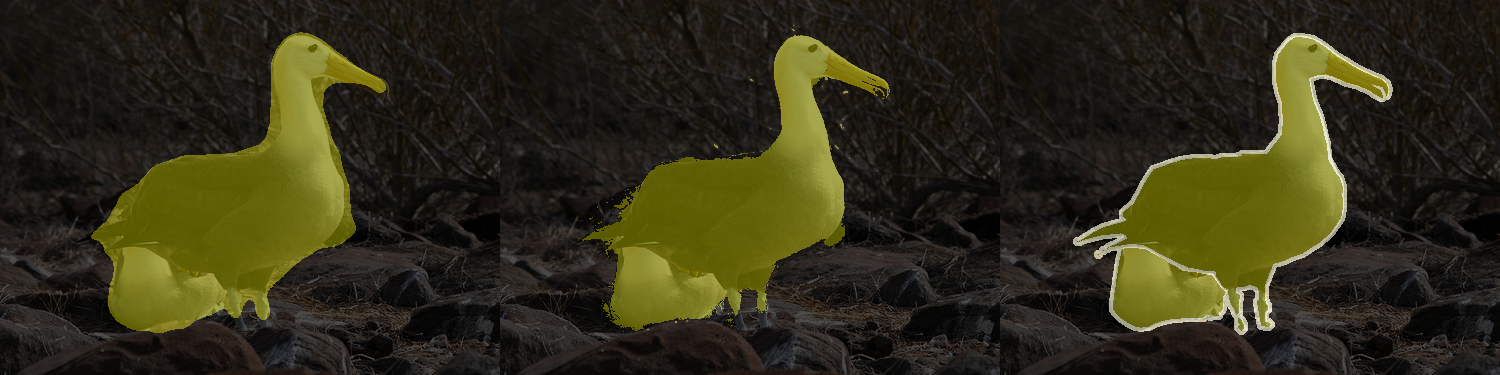}\\
        \includegraphics[width=0.98\linewidth]{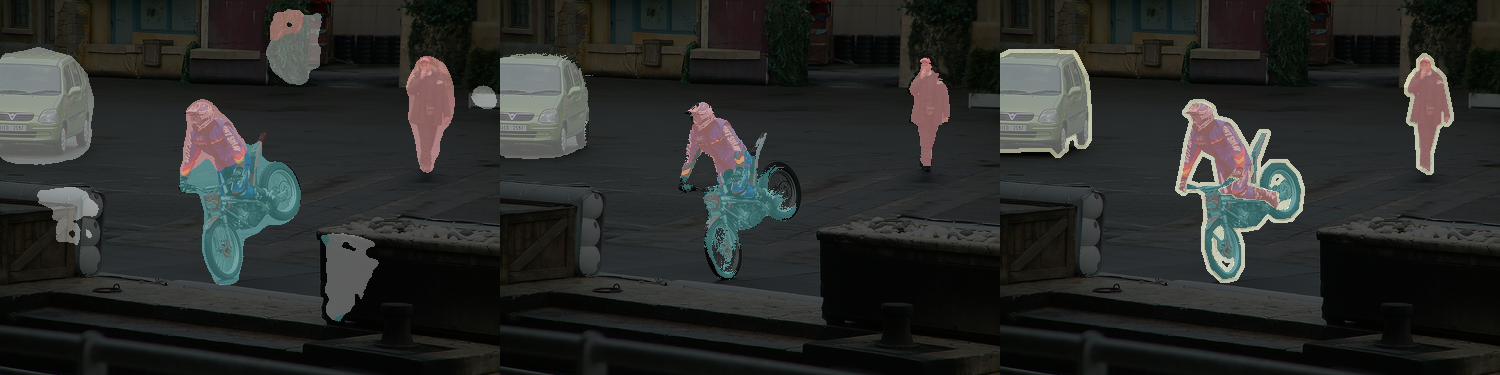}\\
        \includegraphics[width=0.98\linewidth]{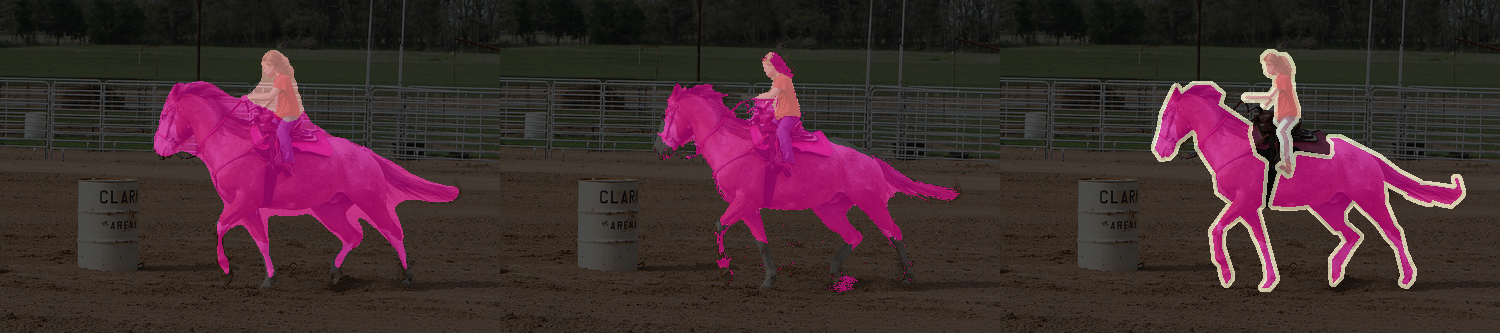}\\

        \includegraphics[width=0.98\linewidth]{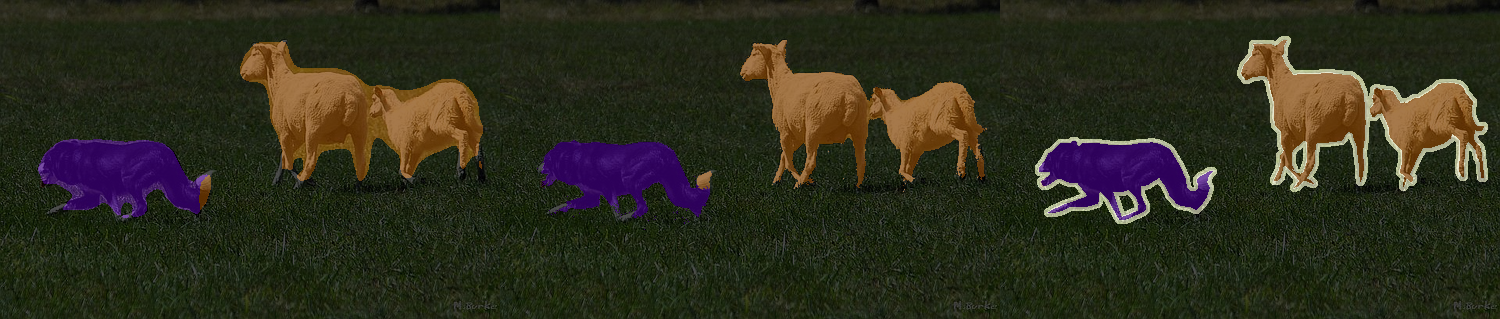}\\
        \includegraphics[width=0.98\linewidth,trim=0 10 0 40,clip]{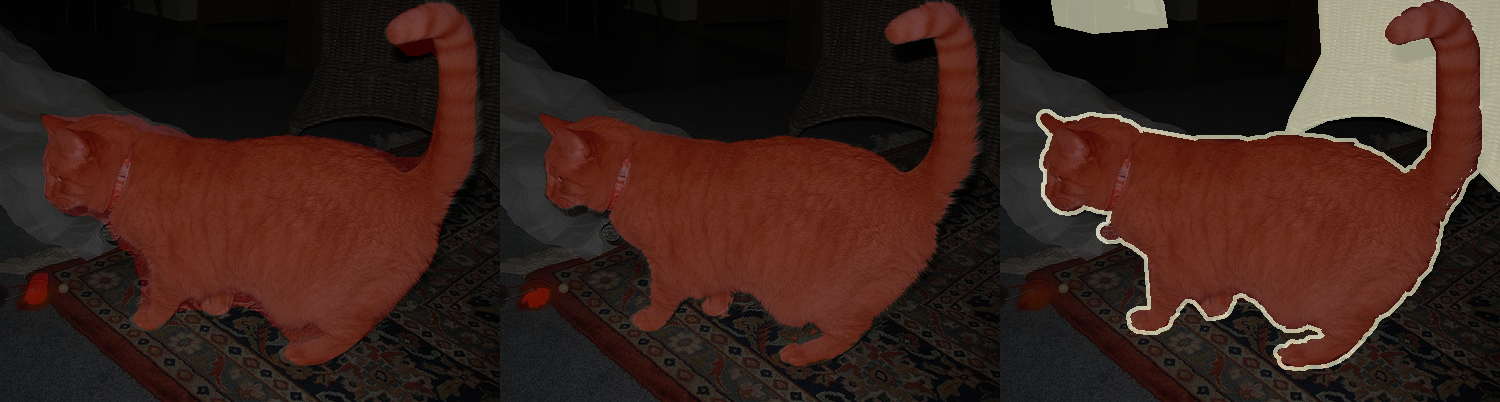}\\
    \end{subfigure}%
    \caption{{Qualitative results on Pascal VOC 2012 \textit{val} set.}
    \textbf{left:} simple cases. \textbf{right:} cases with clutter or incompleteness.
    The SLRNet produces high-quality segmentation results for challenging scenes with varying object sizes and semantic contents.
    }
    \label{fig:qualitative}%
    \vspace{-0.5em}
\end{figure*}

\begin{table}[!t]
  \footnotesize
  \centering
  \begin{tabularx}{\linewidth}{@{}XS[table-format=2.1]@{\hspace{1em}}S[table-format=2.1]@{\hspace{2em}}S[table-format=2.1]@{}}
  \toprule
  Method & CRF & {train(\%)} & {val(\%)} \\
  \midrule
  CAM \tiny{\citep{AhnK18:PSA}} &   & 48.0 & 46.8 \\
  SCE \tiny{\citep{ChangWHPT020:subcat}} & & 50.9 & 49.6 \\
  SEAM \tiny{\citep{WangZKSC20:SEAM}}  &  \checkmark & 56.8 & {--} \\
  \midrule
  CAM+RW \tiny{\citep{AhnK18:PSA}}  &  \checkmark & 59.7 & {--} \\
  SCE+RW \tiny{\citep{ChangWHPT020:subcat}}  &   & 63.4 & {61.2} \\
  \midrule
  1-Stage \tiny{\citep{Araslanov020:SingleStage}}   &   & 64.7 & 63.4 \\
  1-Stage \tiny{\citep{Araslanov020:SingleStage}}  &  \checkmark  &  66.9 & 65.3 \\
  \toprule
  Ours  &             &  67.1 & 66.2 \\
  Ours+CRF  &  \checkmark &  70.3 &  69.3 \\
  \bottomrule
  \end{tabularx}
  \caption{Pseudo-mask quality on the Pascal VOC train and val sets. Here we use image-level labels to filter out false positive errors. }
 \label{tab:psuedo_mask_quality}
\end{table}

\paragraph{Experimental Setup.}
To evaluate the effectiveness of our SLRNet for WSSS, we conduct experiments on the Pascal VOC 2012 dataset~\citep{EveringhamGWWZ10:VOC}, which is a widely-used WSSS benchmark. Following the previous standard practice, we take additional annotations from ~\citep{HariharanABMM11:SBD} to build the augmented training set. In total, 10,582 images are used for training, and 1,449 images are kept for validation.
Note that, only image-level annotations are available during weakly-supervised training.

\paragraph{Pseudo-mask Quality.}
Most of the advanced methods refine the pseudo-masks and distill them into a fully-supervised segmentation network at the last stage.
Although the SLRNet does not need intermediate pseudo-masks for further training,
\Tab\ref{tab:psuedo_mask_quality} compares our pseudo-mask quality with previous state-of-the-arts.
We use image-level ground-truth to filter out false-positive errors following prior practice (\textit{for this experiment only}).
Our method achieves superior performance against improved CAM-generating methods ~\citep{WangZKSC20:SEAM,ChangWHPT020:subcat}, multi-stage CAM-refinement methods ~\citep{AhnK18:PSA}, and single-stage methods ~\citep{Araslanov020:SingleStage}.

\begin{table}[!t]
  \footnotesize
  \begin{tabularx}{\linewidth}{@{}X@{}l@{\hspace{0em}}c@{\hspace{0.5em}}S[table-format=2.1]@{\hspace{0em}}S[table-format=2.1]@{}}
  \toprule
  Method & Backbone & Sup. &  {val(\%)} & {test(\%)} \\
  \midrule
  \multicolumn{5}{@{}l}{\scriptsize \textit{Multi stage (+saliency)}} \\
  \midrule
 SEC \tiny\citep{KolesnikovL16:SEC} & VGG16 & $\mathcal{I},\mathcal{S}$ &   50.7 & 51.7 \\
  MDC \tiny\citep{WeiXSJFH18:mdc} & VGG16 & $\mathcal{I},\mathcal{S}$ & 60.4 & 60.8 \\
  DSRG \tiny\citep{HuangWWLW18:DSRG} & ResNet101 & $\mathcal{I},\mathcal{S}$ &   61.4 & 63.2 \\
  FickleNet \tiny\citep{LeeKLLY19:FickleNet} & ResNet101 & $\mathcal{I},\mathcal{S}$ &   64.9 & 65.3 \\
  CIAN \tiny\citep{FanZTSX20:cian} & ResNet101 & $\mathcal{I},\mathcal{S}$ &   64.3 & 65.3 \\
  MCIS \tiny\citep{SunWDG20:MCIS} & ResNet101 & $\mathcal{I},\mathcal{S}$ & 66.2 & 66.9 \\
  \midrule
  \multicolumn{5}{@{}l}{\scriptsize \textit{Multi stage}} \\
  \midrule
  AffinityNet \tiny\citep{AhnK18:PSA} & ResNet38 & $\mathcal{I}$ &   61.7 & 63.7 \\
  IRN \tiny\citep{AhnCK19:IRN} & ResNet50 & $\mathcal{I}$ & 63.5 & 64.8 \\
  IAL \tiny\citep{WangLMY20:WSSSIAL} & ResNet38 & $\mathcal{I}$ &   64.3 & 65.4 \\
  SSDD \tiny\citep{Shimoda2019:SSDD} & ResNet38 & $\mathcal{I}$ &   64.9 & 65.5 \\
  SEAM \tiny\citep{WangZKSC20:SEAM} & ResNet38 & $\mathcal{I}$ &  64.5 & 65.7 \\
  SCE \tiny\citep{ChangWHPT020:subcat} & ResNet101 & $\mathcal{I}$ & 66.1 & 65.9 \\
  CONTA \tiny\citep{dong_2020:conta} & ResNet38 &  $\mathcal{I}$ & 66.1 & 66.7 \\
  \midrule
  \multicolumn{5}{@{}l}{\scriptsize \textit{Single stage}} \\
  \midrule
  EM-Adapt \tiny\citep{PapandreouCMY15:EM} & VGG16 & $\mathcal{I}$ &  38.2 & 39.6 \\
  MIL-LSE \tiny\citep{PinheiroC15:fromimage} & Overfeat & $\mathcal{I}$ &  42.0 & 40.6 \\
  CRF-RNN \tiny\citep{Zheng15:CRFRNN} & VGG16 & $\mathcal{I}$ &   52.8 & 53.7 \\
  {1-Stage} \tiny\citep{Araslanov020:SingleStage} & ResNet38 & $\mathcal{I}$ & 62.7 & 64.3 \\
  \midrule
  SLRNet (ours) &  \multirow{2}{*}{ResNet38}   & $\mathcal{I}$ &    67.2 & 67.6 \\
  SLRNet distilled$^\ast$ (ours) &   & $\mathcal{I}$ &    69.3 & 69.4 \\
  \bottomrule
  \end{tabularx}
  \caption{Results for WSSS
  on the Pascal VOC validation and test sets.
  Cues used for training are given in the ``Sup.'' column, including image-level labels ($\mathcal{I}$) and saliency detection ($\mathcal{S}$). $\ast$ indicates our \textit{multi-stage} extension: we simply train a DeeplabV3+ network with pseudo-labels generated by SLRNet.
  }
  \label{table:weakly_result}
\end{table}

\paragraph{Experimental Results.}
\Tab\ref{table:weakly_result} compares the SLRNet with a variety of leading single- and multi-stage WSSS methods.
Among them, the single-stage SLRNet achieves the best performance on both val ($67.2\%$) and test ($67.6\%$) sets.
Compared with MCIS~\citep{SunWDG20:MCIS}, the current best-performing multi-stage model with saliency maps, our SLRNet obtains an improvement of $1.0\%$ on the val set.
Compared with SEAM+CONTA~\citep{dong_2020:conta},
that is the current best-performing multi-stage models with WideResNet38 backbone,
our SLRNet achieves an mIoU improvement of $1.1\%$.
Note that the compared multi-stage methods without saliency detection are trained in at least \emph{three} stages, which improve performance at the cost of a considerable increase in model complexity.
Essentially, CONTA~\citep{dong_2020:conta} is a refinement approach that employs additional networks to revise the masks produced by AffinityNet. SEAM~\citep{WangZKSC20:SEAM} and SCE~\citep{ChangWHPT020:subcat} are improved CAM-generating networks which produce the CAMs as AffinityNet's input.
Multi-stage methods improve performance at the cost of a considerable increase in model complexity.
Our SLRNet significantly outperforms previous single-stage models with the help of simple cross-view supervision and the lightweight CVLR.
Besides, although it is trivial to train an additional network for distillation, we still provide a simple distilled result for reference.

\paragraph{Qualitative Analysis.}
\Fig\ref{fig:qualitative} shows some typical qualitative results produced by our SLRNet.
Although only trained with image-level supervision, the SLRNet successfully produces high-quality segmentation results for objects of various sizes in various scenarios.
In the right side of \Fig\ref{fig:qualitative}, we also provide some typical failure cases with clutter or incompleteness, such as interweaving objects and  misleading appearance cues.

\subsection{Experiment II: Learning WSSS from COCO Dataset}

\begin{table}[!t]
  \begin{tabularx}{\linewidth}{@{}Xlc@{}}
  \toprule
  Method & Backbone & {mIoU(\%)}\\
  \midrule
  BFBP  \citep{SalehASPGA16:BFBP}  & VGG16  &  20.4  \\
  SEC \tiny\citep{KolesnikovL16:SEC}   & VGG16  &  22.4  \\
  IAL \citep{WangLMY20:WSSSIAL} & ResNet38  &  27.7  \\
  SEAM$^\ast$ \citep{WangZKSC20:SEAM} & ResNet38  &  31.9  \\
  IRNet$^\ast$ \citep{AhnCK19:IRN} & ResNet50  &  32.6  \\
  SEAM+CONTA \citep{dong_2020:conta} & ResNet38  &  32.8  \\
  IRNet+CONTA \citep{dong_2020:conta} & ResNet50  &  33.4  \\
  \midrule 
  SLRNet (ours) &  {ResNet38}   & {35.0} \\
  \bottomrule
  \end{tabularx}
  \caption{Results for WSSS on COCO validation set. $^\ast$ denotes results provided by ~\cite{dong_2020:conta}.}
  \label{table:COCO_result}
\end{table}

\paragraph{Experimental Setup.}
COCO~\citep{LinMBHPRDZ14:COCO} contains $80$ classes, $80k$, and $40k$ images for training and validation. Although pixel-level labels are provided in the COCO dataset, we only used image-level class labels in the training process.
Note that \textit{we only sample $50\%$ ($40k$) of the training images for training to reduce computational costs.}

\paragraph{Experimental Results.}
\Tab\ref{table:COCO_result} compares our approach and current top-leading WSSS methods with image-level supervision on the COCO dataset. 
We can observe that our method achieves mIoU score of 35.0\% on the val set, outperforming all the competitors.
When compared with IRNet+CONTA~\citep{dong_2020:conta}, the current best-performing method, our approach obtains the improvement of 1.6\% with $50\%$ training samples and simple single-stage training.
Our SLRNet demonstrates a powerful efficiency and efficacy advantage when training on large-scale datasets.
In contrast, most recent approaches, including SEAM~\citep{WangZKSC20:SEAM}, IRNet~\citep{AhnCK19:IRN} and CONTA~\citep{dong_2020:conta}, require training in \textit{three} or \textit{four} stages and search a large number of hyperparameters for each stage.
Moreover, the intermediate results of each stage must be stored on the hard disk which means a huge amount of space and time consumption for large-scale datasets, severely reducing the practicability of WSSS.

\subsection{Experiment III: Performance on the WSSS Track of the L2ID Challenge}
\begin{table}[!t]
    \begin{tabularx}{\linewidth}{@{}l|Xccc@{}}
    \toprule
    Year & Team & Saliency & val(\%) & {test(\%)}\\
    \midrule
    \multirow{2}{*}{LID$_{19}$} & LEAP\_DEXIN& $\checkmark$ & 20.67 & 19.55\\
    & MVN & $\checkmark$ & 40.99& 40.0\\
    \midrule 
    \multirow{3}{*}{LID$_{20}$} & UCU\&SoftServe &  & 39.65 & 37.34 \\
    & VL-task1 & & 40.08 & 37.73  \\
    & CVL & & 46.20 & 45.18   \\
    \midrule
    \multirow{4}{*}{L2ID$_{21}$} &{LEAP Group} & & 40.94& 39.03 \\
    &\scriptsize{NJUST-JDExplore} & & 42.18 & 39.68 \\ 
    &jszx101 & & 50.90 & 49.06 \\
    & SLRNet (ours) & $ $ & {52.30}   & {49.03} \\
    \bottomrule
    \end{tabularx}
    \caption{Experimental results for WSSS on L2ID validation and test set.}
    \label{table:lid_result}
  \end{table}
  
  \paragraph{Experimental Setup.}
    The L2ID challenge dataset~\citep{YunchaoWei2020:lid20} is built upon the object detection track of ImageNet Large Scale Visual Recognition Competition (ILSVRC)~\citep{DengDSLL009:imagenet}, which contains 349,319 images with image-level labels from 200 categories.
    \prelim{
    The evaluation is performed on the validation and test sets, which include 4,690 and 10,000 images, respectively.
    }
    We obtain the pseudo-labels using our single-stage model with  mIoU=52.5\% on the validation set.
    Following the previous practice, we train a standalone Deeplabv3+~\citep{deeplabv3plus2018} network using pseudo-labels generated by our SLRNet.
\paragraph{Experimental Results.}
  \Tab\ref{table:lid_result} lists the final results with the mIoU criterion for WSSS track of L2ID challenge, where the top performing methods are included.
  Our model significantly outperforms the winner of LID 2019, which utilizes saliency maps. In contrast to the champion of LID 2020, the SLRNet is trained in only one cycle to generate pseudo-labels, while the winner integrates the equivariant attention~\citep{WangZKSC20:SEAM} and the co-attention~\citep{SunWDG20:MCIS} to train the classification network, and use the online attention accumulation~\citep{JiangHCCWX19:OAA} to generate object localization maps. Besides, it trains the AffinityNet to refine the pseudo-labels, and leverages the CRF to refine the final segmentation results.
  Our model achieves 52.3\% mIoU on the validation set and 49.03\% mIoU on the test set, respectively, setting a new state-of-the-art on the L2ID challenge through a simple framework.

\subsection{Experiment IV: SSSS with Image-level Labeled Dataset and Pixel-level Labeled Dataset}

\begin{table}[!t]
  \begin{tabularx}{\linewidth}{@{}X@{}l@{\hspace{0.5em}}c@{\hspace{0.5em}}S[table-format=2.1]@{\hspace{0.5em}}S[table-format=2.1]@{}}
  \toprule
  Method & Backbone & Sup. &  {val(\%)} & {test(\%)} \\
  \midrule
  GAIN \tiny\citep{LiWPE018:GAIN} & VGG16 & $\mathcal{I},\mathcal{P},\mathcal{S}$ & 60.5 & 62.1 \\
  DSRG \tiny\citep{HuangWWLW18:DSRG} & VGG16 & $\mathcal{I},\mathcal{P},\mathcal{S}$ &   64.3 & {--} \\
  MDC \tiny\citep{WeiXSJFH18:mdc} & VGG16 & $\mathcal{I},\mathcal{P},\mathcal{S}$ & 65.7 & 67.6 \\
  FickleNet \tiny\citep{LeeKLLY19:FickleNet} & VGG16 & $\mathcal{I},\mathcal{P},\mathcal{S}$ &  65.8 & {--}  \\
  \midrule
  GANSeg  \tiny\citep{SoulySS17:GANSeg}  & VGG16 & $\mathcal{I},\mathcal{P}$ &  65.8 & {--}  \\
  CCT \tiny\citep{OualiHT20:CCT} & ResNet50 & $\mathcal{I},\mathcal{P}$ &  73.2 & {--}  \\
  PseudoSeg \tiny\citep{zou2020:pseudoseg} & ResNet50 & $\mathcal{I},\mathcal{P}$ &  73.8 & {--}  \\
  \midrule
  SLRNet (ours) &  {ResNet38}  & $\mathcal{I},\mathcal{P}$ &    {75.1} &  {75.5} \\
  \bottomrule
  \end{tabularx}
  \caption{Experimental results for semi-supervised semantic segmentation on Pascal VOC validation and test set with $9k$ image-level labeled data $(\mathcal{I})$ and $1.4k$ pixel-level labeled data $(\mathcal{P})$. We also indicate additional saliency supervision $(\mathcal{S})$.}
  \label{table:semi_result}
\end{table}
\paragraph{Experimental Setup.}
To evaluate the proposed method in the semi-supervised setting, we conduct experiments on the Pascal VOC 2012 dataset~\citep{EveringhamGWWZ10:VOC}, which is a standard benchmark of SSSS.
Following the prior practice, we took 1,449 images with pixel-level labels from the VOC training set and an additional 9,133 images with image-level labels from SBD~\citep{HariharanABMM11:SBD} to construct the augmented training set.
Note that the finely labeled and weakly labeled data are mixed and fed to the network in one training cycle.
Intuitively, the finely labeled samples deserve larger weights.
We oversample the finely-labeled data by $5\times$ and multiply their loss by a scaling factor of 2.
We do not use any additional post-processing methods during testing.

\paragraph{Experimental Results.}
\Tab\ref{table:semi_result} lists the comparison results of that our SLRNet to a variety of state-of-the-art methods on the Pascal VOC dataset, where our SLRNet is trained on only 13.8\% of images with pixel-level annotations. It achieves an mIoU of 75.1\%, which significantly surpasses the WSSS-based methods~\citep{HuangWWLW18:DSRG,WeiXSJFH18:mdc,LeeKLLY19:FickleNet}, GAN-based methods~\citep{SoulySS17:GANSeg}, and consistency-based methods~\citep{OualiHT20:CCT,zou2020:pseudoseg}.

\subsection{Experiment V: SSSS with Pixel-level Labeled Dataset and Unlabeled Dataset}
\begin{table}[!h]
  \begin{tabularx}{\linewidth}{@{}Xlc@{}}
  \toprule
  Method & Backbone & {mIoU (\%)} \\
  \midrule
  GANSeg  \tiny\citep{SoulySS17:GANSeg}  & VGG16  &  64.1  \\
  AdvSemSeg \tiny\citep{HungTLL018:AdvSemSeg}   & ResNet101  &  68.4  \\
  CCT \tiny\citep{OualiHT20:CCT} & ResNet50  &  69.4  \\
  PseudoSeg \tiny\citep{zou2020:pseudoseg} & ResNet101  &  72.3  \\
  \midrule 
  \multirow{2}{*}{SLRNet (ours)} &  {ResNet38}   & {72.4} \\
   &  {ResNet101}  & {72.9} \\
  \bottomrule
  \end{tabularx}
  \caption{Experimental results for semi-supervised semantic segmentation on the Pascal VOC validation set with $1.4k$ pixel-level labeled data $(\mathcal{P})$ and $9k$ unlabeled data.}
  \label{table:semi_result_wo_cls}
\end{table}

\paragraph{Experimental Setup.}
We also conduct experiments on the Pascal VOC 2012 dataset~\citep{EveringhamGWWZ10:VOC} using 1,449 pixel-level labeled images and an additional 9,133 unlabeled images.

\paragraph{Experimental Results.}
\Tab\ref{table:semi_result_wo_cls} compares the performance of our SLRNet with previous state-of-the-art SSSS methods.
It is worth pointing out that we exactly use the same network and hyperparameters as in the image-level WSSS setting. Notwithstanding, the SLRNet still outperforms all the other dedicated SSSS models, illustrating the good versatility and generality of our approach.
\subsection{Ablation Study}
\label{sec:ablation}
We conduct ablation experiments on the Pascal VOC dataset for WSSS settings. 
To demonstrate the improvement source of our SLRNet, we use \textit{mean false discovery rate} (mFDR) and \textit{mean false negative rate} (mFNR) to indicate the semantic fidelity and completeness respectively, which are defined as 
\begin{equation}
  mFDR=\frac{1}{C}\sum_{c=1}^{C}{\frac{FP_c}{TP_c+FP_c}},
\end{equation}
and
\begin{equation}
  mFNR=\frac{1}{C}\sum_{c=1}^{C}{\frac{FN_c}{TP_c+FN_c}},
\end{equation}
where $TP_c$, $FP_c$, and $FN_c$ denote the number of true positive, false positive and false negative predictions of class $c$ respectively.

\begin{table}[t]
    \centering
    \begin{tabular}{l|lcc|r}
       \toprule
       {\#views}&{scales} & {color} & {flip} & {mIoU(\%)} \\
       \midrule
       1 & {$\left(1.0\right)$} & \checkmark & \checkmark & {60.88} \\
       \midrule
       \multirow{4}{*}{2}
         & {$\left(0.5,~1.0\right)$} & \checkmark & \checkmark & \textbf{64.07} \\
         & {$\left(1.0,~1.0\right)$} & \checkmark & \checkmark & {61.53} \\
         & {$\left(0.5,~1.0\right)$} & \checkmark &  & {63.89} \\
         & {$\left(0.5,~1.0\right)$} &  &  & {63.99} \\
       \midrule
       3 & {$\left(0.5,~1.0,~1.5\right)$} &  &  & {63.90} \\
       \bottomrule
    \end{tabular}
    \caption{{Evaluation on different compositions of data augmentation applied on multiple branches. Here, \checkmark means that multiple views use the different random transformations (e.g. for color distortion, randomly different distortion strength for different views). No CRF or any other post-processing is used.}}
 \label{tab:augmentation}
 \vspace{-1em}
 \end{table}

\paragraph{Data Augmentation.}
 To understand the effect of individual data augmentation for our {\name}, we consider several geometric and appearance augmentations in~\citep{ChenK0H20:SimCLR}.
 Furthermore, we pay more attention to the \textit{invertible} and \textit{differentiable} geometric transformations, such as resizing and flipping.
 First, the images are randomly cropped to $321\times321$. Then, we apply target transformations to different branches.
 We study the compositions of three kinds of transformations: re-scaling with fixed rates, random horizontal flip and random color distortion (\textit{e.g.} brightness, contrast, saturation, and hue).
Stronger color distortion cannot improve even hurts performance under supervised settings~\citep{ChenK0H20:SimCLR,lee2021improving}.
Therefore, we set the maximum strength of color distortion to 0.3 for brightness, contrast, and saturation, and 0.1 for hue component.

\Tab\ref{tab:augmentation} lists evaluation results on the the Pascal VOC val set under different compositions of transformations.
We observe that \textit{combination of three different augmentations produces the best performance} ($64.07\%$).
When composing more augmentations, cross-view supervision is expected to become much stronger.
We also note that \textit{re-scaling contributes a significantly greater improvement than other augmentations.}
There is a significant mIoU drop ($2.54\%$) without re-scaling.
In contrast, using the same color distortion and flipping for different views leads to a slight mIoU drop ($0.08\%$).
The combination of different color distortion and flipping only achieves a minor improvement ($0.65\%$) compared with single-view baseline.
Furthermore, it is worth pointing out that although adding more views has higher complexity, this cannot improve the performance.
This indicates that the crucial improvement sources are our cross-view design and the LR reconstruction, which require simple augmentations to provide perturbations.

 \begin{figure}[t!]
    \centering
    \begin{subfigure}{0.6\linewidth}
      \includegraphics[width=\linewidth]{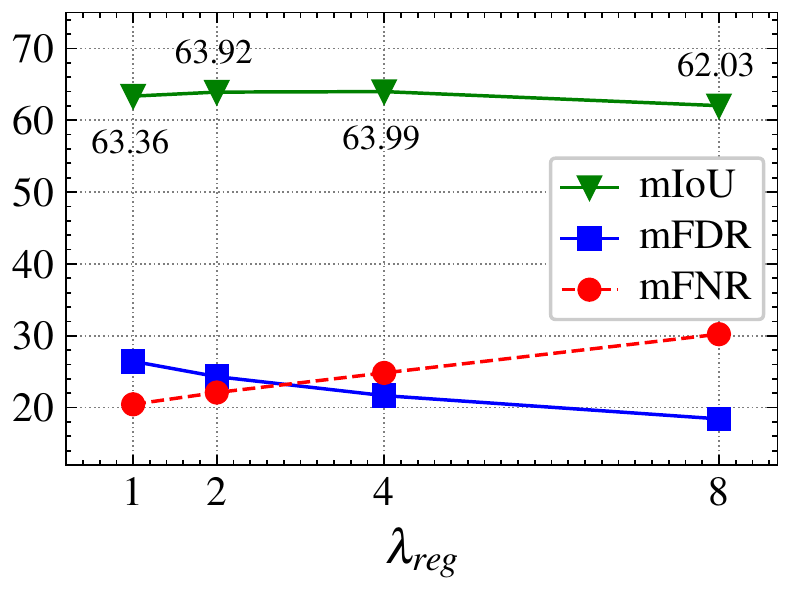}
    \end{subfigure}
    \begin{subfigure}{0.38\linewidth}
        \centering
        \begin{tabular}{l | c}
            Sup.  & mIoU \\
            \midrule
            CPS & 20.78 \\
            KL Div. & 47.13 \\
            \midrule
            MVMC & 63.99 \\
        \end{tabular}
    \end{subfigure}
    \caption{
        \prelim{
        Ablation study on cross-view supervision. \textbf{Left plot:} segmentation performance with varied explicit cross-view supervision strength ($\lambda_{reg}$). \textbf{Right table:} comparison with other cross-view supervision, including Kullback–Leibler divergence (KL Div.) and cross pseudo supervision (CPS) ~\citep{Chen2021:CrossPseudo}.
        }
    }
    \label{fig:ablation_mvmc}
\end{figure}

\paragraph{Cross-view Supervision.}
Cross-view supervision aims to mitigate the compounding effect of self-supervision errors by introducing additional regularization.
We adjust the loss coefficient $\lambda_\text{reg}$ that explicitly controls the strength of cross-view supervision.
As shown in \Fig\ref{fig:ablation_mvmc}, we observe that \textit{cross-view supervision mainly improves the segmentation quality by reducing $mFDR$, \textit{i.e.}, preventing the accumulation of false positives in self-supervision to improve the semantic fidelity}.
This improvement is maximized with $\lambda_\text{reg}=4$ in our experiments. (For clarification, the scale of  $\mathcal{L}_\text{reg}$ is much smaller than  $\mathcal{L}_\text{seg}$.)
It is worth pointing out that \textit{higher strength of cross-view supervision increases $mFNR$, \textit{i.e.}, hurts semantic completeness.}
Collapsing solutions, where the predictions are all ``background'', will occur when explicit cross-view supervision dominates ~\citep{Xinlei:SimSiam,CaronMMGBJ20:SWaV}.
Therefore in the task with limited supervision, choosing appropriate cross-view strength is the crux of performance improvement.

\prelim{
To further validate efficacy of the proposed implicit cross-view supervision, we also compare the MVMC with other cross-view supervision mechanisms.
As shown in \Fig\ref{fig:ablation_mvmc} (right), we substitute the cross pseudo supervision~\citep{Chen2021:CrossPseudo} or Kullback–Leibler divergence for MVMC, achieving much worse results.
Generally, these explicit consistency methods are applied to SSSS that require pixel-level supervision to avoid collapsing solutions~\citep{Xinlei:SimSiam}.
}

\begin{table}
        \centering
        \begin{tabularx}{\linewidth}{ X | c c | c }
            \toprule
            Method & \#Iter. & Grad.  & mIoU(\%) \\
            \midrule
            CVLR & 1 & yes &   \textbf{63.99}\\
            - Separate dictionary $\mathbf{D}$ & 1 & yes &   61.24\\
            - w/o latent space reg. & 3 & yes & 62.83\\
            - w/o CVLR & {--} & {--} &   61.11\\
            \midrule
            EMAU~\small\citep{Lixia19:EMANET} & 3 & no &   61.76\\
            Ham.~\small\citep{Zheng:ham} & 6 & 1 step &  59.75  \\
            \bottomrule
        \end{tabularx}
        \caption{
        \prelim{Ablation on shared dictionary $\mathbf{D}$ and latent space regularization. We also provide comparisons between CVLR and other LR modules in terms of iterations (\#Iter.) and back-propagation gradients (Grad.).}
        }
       \label{tab:ablation_cvlr_parts}
\end{table}

 \paragraph{The CVLR Module.}
\prelim{
To study the effectiveness of each part in the CVLR, we conduct thorough ablation experiments on it.
Firstly, as listed in \Tab\ref{tab:ablation_cvlr_parts}, substituting the shared dictionary matrix with separate ones causes the most severe decay in performance, attesting the significance of the cross-view LR mechanism.
Second, the latent space regularization also contributes considerable performance improvement, as it further enforces the semantic stability and consistency of the elements in the shared dictionary and reduces optimization steps.
Furthermore, \Tab\ref{tab:ablation_cvlr_parts} compares the CVLR with the EMAU~\citep{Lixia19:EMANET} and the Hamburger~\citep{Zheng:ham} modules that learn single-view LR representations with specific designs, demonstrating  advantages of the proposed CVLR.
}

\begin{figure}[t]
    \def\svgwidth{\linewidth}
    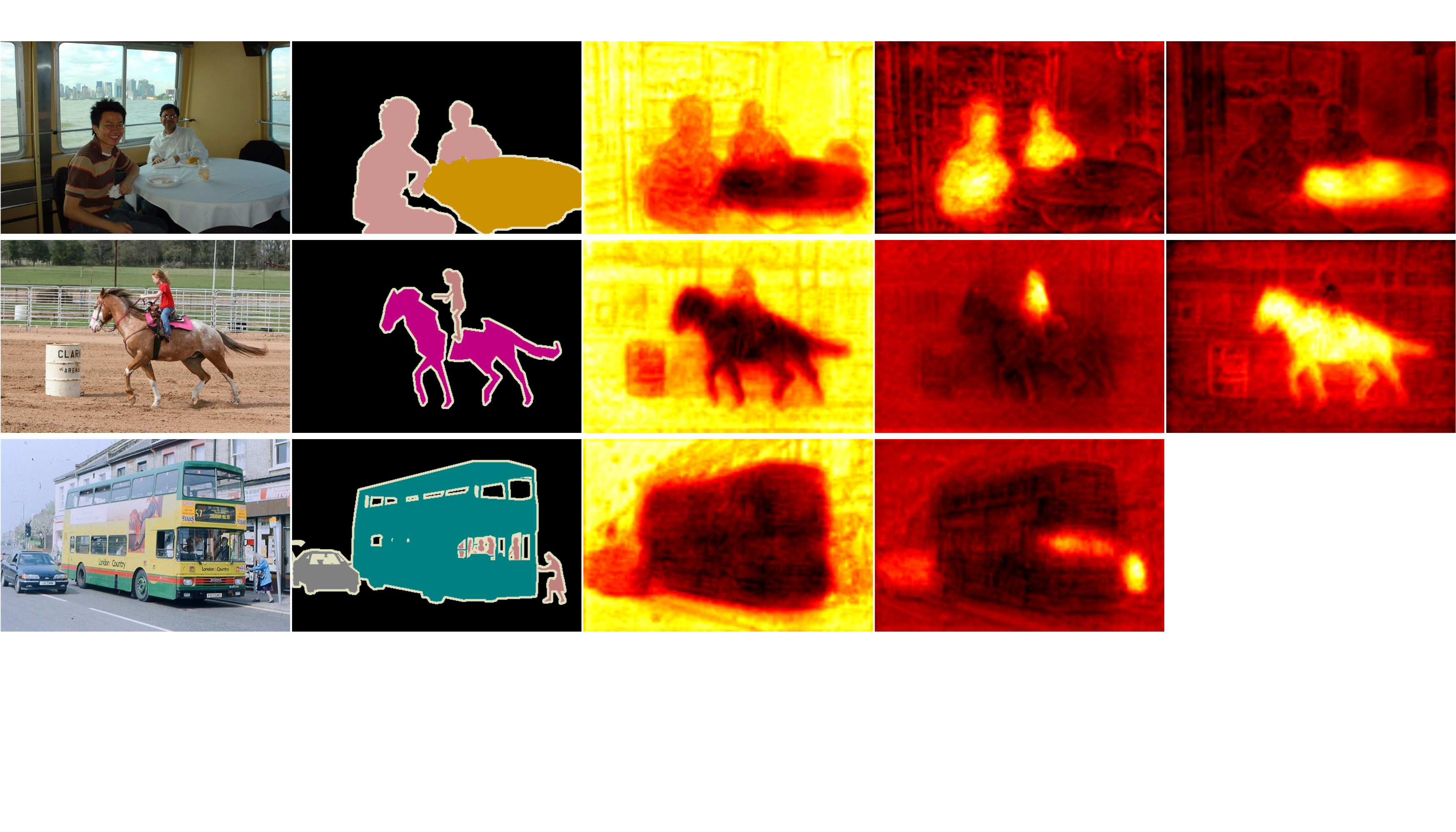
    \vspace{-1.5em}
    \caption{
        Visualization of code $\mathbf{C}_{k,\cdot}$ for specific item in dictionary $\mathbf{D}$.
    }
    \label{fig:assignment_vis}
\end{figure}

To understand and verify the behavior of CVLR, as in \Fig~\ref{fig:framework} (right), we visualize the feature map before and after the module. The CVLR emphasizes and refines the relevant features from complementary views through reducing ambiguous activation and completing regions.
Meanwhile, the proposed latent space regularization allows a close look at the collective matrix factorization, making the factor matrices well interpretable.
\Fig~\ref{fig:assignment_vis} visualizes the optimal factor matrix $\mathbf{C}$ (single view is selected) that represents the corresponding probabilities of all pixels to the selected $k$-th element in the dictionary.
The code matrix of specific categories completely highlights the corresponding semantic regions and finely delineates the boundaries.

\begin{figure}
    \centering
    \begin{subfigure}{0.48\linewidth}
          \includegraphics[width=\textwidth]{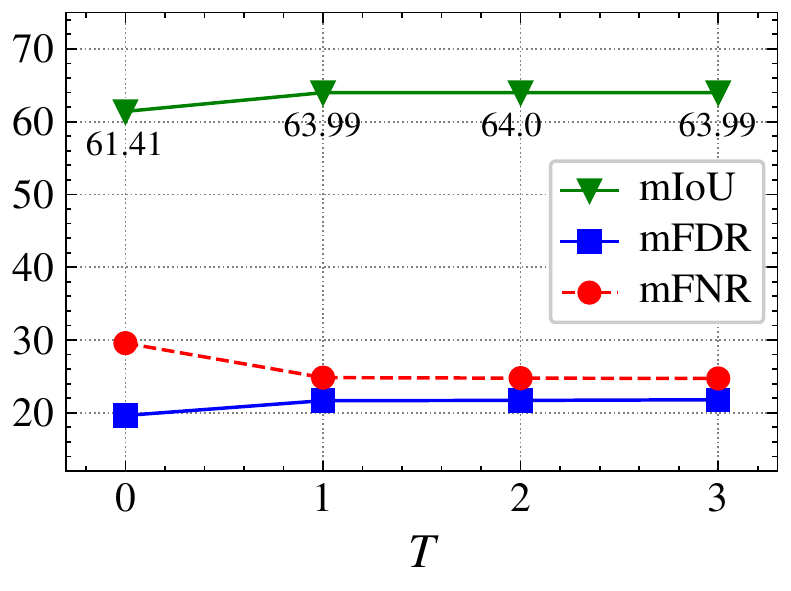}
      \vspace{-1em}
    \end{subfigure}
    \begin{subfigure}{0.48\linewidth}
      \includegraphics[width=\textwidth]{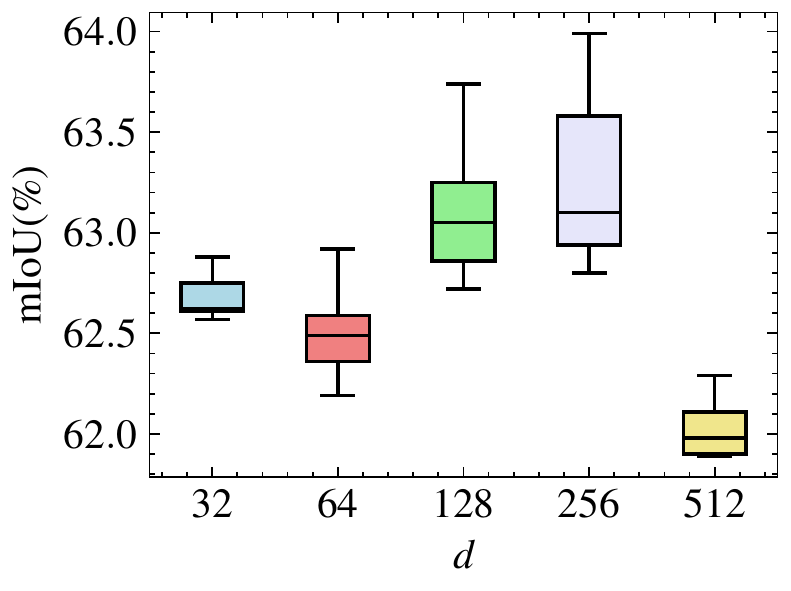}
      \vspace{-1em}
    \end{subfigure}
    \caption{
        \prelim{
        Ablation on the iteration number $T$ of CVLR and dimension $d$ of the dictionary matrix.
        } 
    }
    \label{fig:ablation_cvlr_params}
\end{figure}

\begin{figure}[!t]
    \def\svgwidth{\linewidth}
    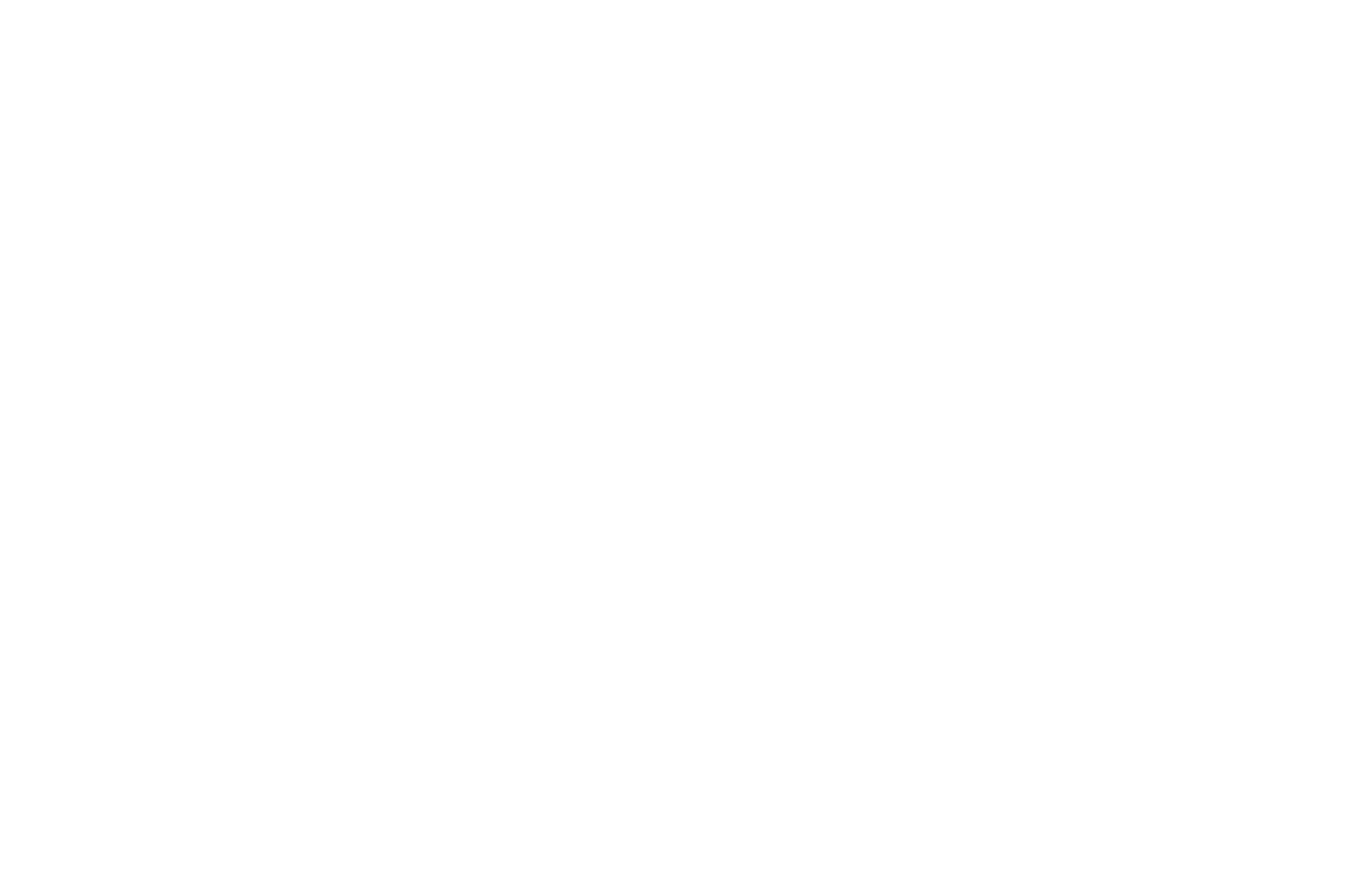
    \vspace{-1.5em}
    \caption{
        Visualization of factor $\mathbf{C}_{k,\cdot}$ with different iteration number $T$. The first two rows are from the early training step (randomly selected, epoch=7) and the last two rows are from the last training step (epoch=20).
    }
    \label{fig:iterations_assignment_vis}
    \vspace{1em}
\end{figure}

Another experiment is conducted to explore the impact of hyper-parameters in the CVLR module, including optimization step  $T$ and latent space dimension $d$.
As shown in \Fig~\ref{fig:ablation_cvlr_params} (left), we observe that \textit{the iteration steps improve the segmentation performance mainly by reducing $mFNR$, i.e., improving the semantic completeness}. 
We also find that single-step optimization is enough and more iterations cannot improve performance.
To empirically analyze this observation, \Fig\ref{fig:iterations_assignment_vis} visualizes the factor matrix $\mathbf{C}_{k,\cdot}$ with different number of iterations at different training step.
More iterations yield evident effects in the early epochs, while becoming insignificant as the network is converged.
Inspired by SimSiam~\citep{Xinlei:SimSiam}, we conjecture that the multi-step alternation (inner loop) is optional because the SGD steps provide much more  outer loops and the optimization trajectory of the collective MF is memorized by the network parameters.
\prelim{
Furthermore, as illustrated in \Fig\ref{fig:ablation_cvlr_params} (right), either too small or too large latent space dimension will negatively impact the segmentation performance and $d=256$ is the optimal choice in our experiments.
}

\section{Conclusion}
This paper has presented a simple yet effective SLRNet for single-stage WSSS and SSSS under online pseudo-label supervision.
To overcome the compounding effect of self-supervision errors, we have developed a Siamese network based architecture that makes full use of cross-view supervision and the LR property of the features.
Specifically, we have designed the MVMC that aids with explicit cross-view consistency to provide a flexible cross-view supervision solution.
Then, we have built a lightweight CVLR module that can be readily integrated into the network for end-to-end training.
The CVLR learns an interpretable global cross-view LR representation, which complements cross-view supervision to improve semantic completeness while ensuring semantic fidelity.
Despite its simplicity, extensive evaluations have demonstrated that the proposed SLRNet has achieved favorable performance superior to that of both leading single- and multi-stage WSSS and SSSS methods in terms of complexity and accuracy.
Our future work is to extend the proposed single-stage model to other label-efficient tasks without the need of considerable training cycles and post-processing techniques.

\begin{acknowledgements}
This work was partially supported by the National Key Research and Development Program of China under Grant 2019YFB2101904, the National Natural Science Foundation of China under Grants 61732011, 61876127, 61876088 and 61925602.
\end{acknowledgements}

\bibliographystyle{spbasic}
\bibliography{reference}

\end{document}

%% file: figures/framework/overall2.tex

\begingroup%
  \makeatletter%
  \providecommand\color[2][]{%
    \errmessage{(Inkscape) Color is used for the text in Inkscape, but the package 'color.sty' is not loaded}%
    \renewcommand\color[2][]{}%
  }%
  \providecommand\transparent[1]{%
    \errmessage{(Inkscape) Transparency is used (non-zero) for the text in Inkscape, but the package 'transparent.sty' is not loaded}%
    \renewcommand\transparent[1]{}%
  }%
  \providecommand\rotatebox[2]{#2}%
  \newcommand*\fsize{\dimexpr\f@size pt\relax}%
  \newcommand*\lineheight[1]{\fontsize{\fsize}{#1\fsize}\selectfont}%
  \ifx\svgwidth\undefined%
    \setlength{\unitlength}{648.99998474bp}%
    \ifx\svgscale\undefined%
      \relax%
    \else%
      \setlength{\unitlength}{\unitlength * \real{\svgscale}}%
    \fi%
  \else%
    \setlength{\unitlength}{\svgwidth}%
  \fi%
  \global\let\svgwidth\undefined%
  \global\let\svgscale\undefined%
  \makeatother%
  \begin{picture}(1,0.33127889)%
    \lineheight{1}%
    \setlength\tabcolsep{0pt}%

    \put(0,0){\includegraphics[width=\unitlength,page=1]{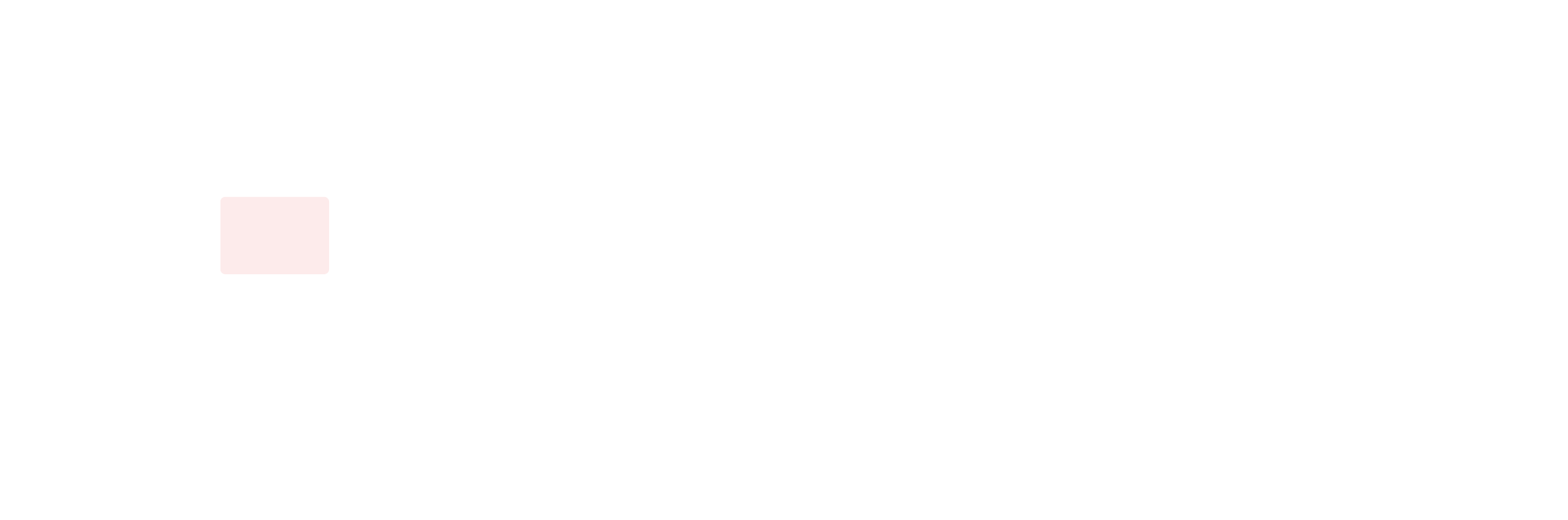}}%
    \put(0,0){\includegraphics[width=\unitlength,page=2]{figures/framework/overall.pdf}}%
    \put(0,0){\includegraphics[width=\unitlength,page=3]{figures/framework/overall.pdf}}%
    \put(0,0){\includegraphics[width=\unitlength,page=4]{figures/framework/overall.pdf}}%
    \put(0,0){\includegraphics[width=\unitlength,page=5]{figures/framework/overall.pdf}}%
    \put(0,0){\includegraphics[width=\unitlength,page=6]{figures/framework/overall.pdf}}%
    \put(0,0){\includegraphics[width=\unitlength,page=7]{figures/framework/overall.pdf}}%
    \put(0,0){\includegraphics[width=\unitlength,page=8]{figures/framework/overall.pdf}}%
    \put(0,0){\includegraphics[width=\unitlength,page=9]{figures/framework/overall.pdf}}%
    \put(0,0){\includegraphics[width=\unitlength,page=10]{figures/framework/overall.pdf}}%
    \put(0,0){\includegraphics[width=\unitlength,page=11]{figures/framework/overall.pdf}}%
    \put(0,0){\includegraphics[width=\unitlength,page=12]{figures/framework/overall.pdf}}%
    \put(0,0){\includegraphics[width=\unitlength,page=13]{figures/framework/overall.pdf}}%
    \put(0,0){\includegraphics[width=\unitlength,page=14]{figures/framework/overall.pdf}}%
    \put(0,0){\includegraphics[width=\unitlength,page=15]{figures/framework/overall.pdf}}%
    \put(0,0){\includegraphics[width=\unitlength,page=16]{figures/framework/overall.pdf}}%
    \put(0,0){\includegraphics[width=\unitlength,page=17]{figures/framework/overall.pdf}}%
    \put(0,0){\includegraphics[width=\unitlength,page=18]{figures/framework/overall.pdf}}%
    \put(0,0){\includegraphics[width=\unitlength,page=19]{figures/framework/overall.pdf}}%
    \put(0,0){\includegraphics[width=\unitlength,page=20]{figures/framework/overall.pdf}}%
    \put(0,0){\includegraphics[width=\unitlength,page=21]{figures/framework/overall.pdf}}%
    \put(0,0){\includegraphics[width=\unitlength,page=22]{figures/framework/overall.pdf}}%
    \put(0,0){\includegraphics[width=\unitlength,page=23]{figures/framework/overall.pdf}}%

    \put(0.15,0.17495838){\color[rgb]{0,0,0}\makebox(0,0)[lt]{\lineheight{1.25}\smash{\begin{tabular}[t]{l}CVLR\end{tabular}}}}%
    \put(0.11,0.2663667){\color[rgb]{0,0,0}\makebox(0,0)[lt]{\lineheight{1.25}\smash{\begin{tabular}[t]{l}Enc\end{tabular}}}}%

    \put(0.2758,0.17514944){\color[rgb]{0,0,0}\makebox(0,0)[lt]{\lineheight{1.25}\smash{\begin{tabular}[t]{l}MVMC\end{tabular}}}}%

    \put(0.215,0.26875499){\color[rgb]{0,0,0}\makebox(0,0)[lt]{\lineheight{1.25}\smash{\begin{tabular}[t]{l}Dec\end{tabular}}}}%

    \put(0.05,0.2301402){\color[rgb]{0,0,0}\makebox(0,0)[lt]{\lineheight{1.25}\smash{\begin{tabular}[t]{l}$\mathcal{T}^{(v)}$\end{tabular}}}}%
    \put(0.05,0.11728134){\color[rgb]{0,0,0}\makebox(0,0)[lt]{\lineheight{1.25}\smash{\begin{tabular}[t]{l}$\mathcal{T}^{(u)}$\end{tabular}}}}%

    \put(0.11,0.08269937){\color[rgb]{0,0,0}\makebox(0,0)[lt]{\lineheight{1.25}\smash{\begin{tabular}[t]{l}Enc\end{tabular}}}}%

    \put(0.215,0.08346978){\color[rgb]{0,0,0}\makebox(0,0)[lt]{\lineheight{1.25}\smash{\begin{tabular}[t]{l}Dec\end{tabular}}}}%

    \put(0.435,0.17615099){\color[rgb]{0,0,0}\makebox(0,0)[lt]{\lineheight{1.25}\smash{\begin{tabular}[t]{l}$\mathcal{L}_{\text{reg}}$\end{tabular}}}}%

    \put(0.407,0.30115561){\color[rgb]{0,0,0}\makebox(0,0)[lt]{\lineheight{1.25}\smash{\begin{tabular}[t]{l}$\mathcal{L}_{\text{cls}}$\end{tabular}}}}%

    \put(0.37,0.23283666){\color[rgb]{0,0,0}\makebox(0,0)[lt]{\lineheight{1.25}\smash{\begin{tabular}[t]{l}$\mathcal{L}_{\text{seg}}$\end{tabular}}}}%

    \put(0.37,0.11729105){\color[rgb]{0,0,0}\makebox(0,0)[lt]{\lineheight{1.25}\smash{\begin{tabular}[t]{l}$\mathcal{L}_{\text{seg}}$\end{tabular}}}}%

    \put(0.407,0.04661015){\color[rgb]{0,0,0}\makebox(0,0)[lt]{\lineheight{1.25}\smash{\begin{tabular}[t]{l}$\mathcal{L}_{\text{cls}}$\end{tabular}}}}%

    \put(0.258,0.2278459){\color[rgb]{0,0,0}\makebox(0,0)[lt]{\lineheight{1.25}\smash{\begin{tabular}[t]{l}$\mathcal{T}^{(v)\prime}$\end{tabular}}}}%
    \put(0.258,0.11728134){\color[rgb]{0,0,0}\makebox(0,0)[lt]{\lineheight{1.25}\smash{\begin{tabular}[t]{l}$\mathcal{T}^{(u)\prime}$\end{tabular}}}}%

    \put(0.04534807,0.0086467){\color[rgb]{0,0,0}\makebox(0,0)[lt]{\lineheight{1.25}\smash{\begin{tabular}[t]{l}Share parameters\end{tabular}}}}%
    \put(0.52,0.31311709){\color[rgb]{0,0,0}\makebox(0,0)[lt]{\lineheight{1.25}\smash{\begin{tabular}[t]{l}CVLR Module\end{tabular}}}}%
    \put(0.0,0.31311709){\color[rgb]{0,0,0}\makebox(0,0)[lt]{\lineheight{1.25}\smash{\begin{tabular}[t]{l}Framework\end{tabular}}}}%

    \put(0.74,0.26464251){\color[rgb]{0,0,0}\makebox(0,0)[lt]{\lineheight{1.25}\smash{\begin{tabular}[t]{l}$C^{(v)}$\end{tabular}}}}%

    \put(0.8,0.26464251){\color[rgb]{0,0,0}\makebox(0,0)[lt]{\lineheight{1.25}\smash{\begin{tabular}[t]{l}$C^{(u)}$\end{tabular}}}}%

    \put(0.89,0.26564251){\color[rgb]{0,0,0}\makebox(0,0)[lt]{\lineheight{1.25}\smash{\begin{tabular}[t]{l}$E^{(v)}$\end{tabular}}}}%

    \put(0.945,0.26564251){\color[rgb]{0,0,0}\makebox(0,0)[lt]{\lineheight{1.25}\smash{\begin{tabular}[t]{l}$E^{(u)}$\end{tabular}}}}%

    \put(0.674,0.26564251){\color[rgb]{0,0,0}\makebox(0,0)[lt]{\lineheight{1.25}\smash{\begin{tabular}[t]{l}$D$\end{tabular}}}}%

    \put(0.55,0.26564251){\color[rgb]{0,0,0}\makebox(0,0)[lt]{\lineheight{1.25}\smash{\begin{tabular}[t]{l}$X^{(v)}$\end{tabular}}}}%
    \put(0.61,0.26564251){\color[rgb]{0,0,0}\makebox(0,0)[lt]{\lineheight{1.25}\smash{\begin{tabular}[t]{l}$X^{(u)}$\end{tabular}}}}%

    \put(0.84548228,0.26548843){\color[rgb]{0,0,0}\makebox(0,0)[lt]{\lineheight{1.25}\smash{\begin{tabular}[t]{l}+\end{tabular}}}}%

    \put(0.70008783,0.26564251){\color[rgb]{0,0,0}\makebox(0,0)[lt]{\lineheight{1.25}\smash{\begin{tabular}[t]{l}$\times$\end{tabular}}}}%

    \put(0.46,0.30192602){\color[rgb]{0,0,0}\makebox(0,0)[lt]{\lineheight{1.25}\smash{\begin{tabular}[t]{l}$y$\end{tabular}}}}%
    \put(0.46,0.04738057){\color[rgb]{0,0,0}\makebox(0,0)[lt]{\lineheight{1.25}\smash{\begin{tabular}[t]{l}$y$\end{tabular}}}}%

    \put(0.263,0.0086467){\color[rgb]{0,0,0}\makebox(0,0)[lt]{\lineheight{1.25}\smash{\begin{tabular}[t]{l} Optional\end{tabular}}}}%

    \put(0.72,0.138){\color[rgb]{0,0,0}\makebox(0,0)[lt]{\lineheight{1.25}\smash{\begin{tabular}[t]{c}Collective\\MF\end{tabular}}}}%
    
    
    \put(0.71284939,0.027){\color[rgb]{0,0,0}\makebox(0,0)[lt]{\lineheight{1.25}\smash{\begin{tabular}[t]{l}Initialize\end{tabular}}}}%

    \put(0.66072573,0.08016901){\color[rgb]{0,0,0}\makebox(0,0)[lt]{\lineheight{1.25}\smash{\begin{tabular}[t]{l}\small{Linear}\\ \small{projector}\end{tabular}}}}%
    \put(0.81026656,0.08016901){\color[rgb]{0,0,0}\makebox(0,0)[lt]{\lineheight{1.25}\smash{\begin{tabular}[t]{l} \small{Linear} \\   \small{projector} \end{tabular}}}}%
    \put(0.5645601,0.01983756){\color[rgb]{0,0,0}\makebox(0,0)[lt]{\lineheight{1.25}\smash{\begin{tabular}[t]{l}Auxiliary\\ Predictor \end{tabular}}}}%
    \put(0.53,0.23092756){\color[rgb]{0,0,0}\makebox(0,0)[lt]{\lineheight{1.25}\smash{\begin{tabular}[t]{l}${d\times h^{(v)} w^{(v)}}$\end{tabular}}}}%
    \put(0.716,0.23092756){\color[rgb]{0,0,0}\makebox(0,0)[lt]{\lineheight{1.25}\smash{\begin{tabular}[t]{l}${k\times h^{(v)} w^{(v)}}$\end{tabular}}}}%
    \put(0.66,0.23092756){\color[rgb]{0,0,0}\makebox(0,0)[lt]{\lineheight{1.25}\smash{\begin{tabular}[t]{l}${d\times k}$\end{tabular}}}}%
    \put(0.86,0.23092756){\color[rgb]{0,0,0}\makebox(0,0)[lt]{\lineheight{1.25}\smash{\begin{tabular}[t]{l}${d\times h^{(v)} w^{(v)}}$\end{tabular}}}}%
    \put(0.65600462,0.26545761){\color[rgb]{0,0,0}\makebox(0,0)[lt]{\lineheight{1.25}\smash{\begin{tabular}[t]{l}=\end{tabular}}}}%

    \put(0.53043605,0.19546686){\color[rgb]{0,0,0}\makebox(0,0)[lt]{\lineheight{1.25}\smash{\begin{tabular}[t]{l}Inputs\end{tabular}}}}%
    \put(0.883,0.19546686){\color[rgb]{0,0,0}\makebox(0,0)[lt]{\lineheight{1.25}\smash{\begin{tabular}[t]{l}Outputs\end{tabular}}}}%

  \end{picture}%
\endgroup%

%% file: figures/assignments/supp_assignments.pdf_tex
\begingroup%
  \makeatletter%
  \providecommand\color[2][]{%
    \errmessage{(Inkscape) Color is used for the text in Inkscape, but the package 'color.sty' is not loaded}%
    \renewcommand\color[2][]{}%
  }%
  \providecommand\transparent[1]{%
    \errmessage{(Inkscape) Transparency is used (non-zero) for the text in Inkscape, but the package 'transparent.sty' is not loaded}%
    \renewcommand\transparent[1]{}%
  }%
  \providecommand\rotatebox[2]{#2}%
  \newcommand*\fsize{\dimexpr\f@size pt\relax}%
  \newcommand*\lineheight[1]{\fontsize{\fsize}{#1\fsize}\selectfont}%
  \ifx\svgwidth\undefined%
    \setlength{\unitlength}{1005.99993896bp}%
    \ifx\svgscale\undefined%
      \relax%
    \else%
      \setlength{\unitlength}{\unitlength * \real{\svgscale}}%
    \fi%
  \else%
    \setlength{\unitlength}{\svgwidth}%
  \fi%
  \global\let\svgwidth\undefined%
  \global\let\svgscale\undefined%
  \makeatother%
  \begin{picture}(1,0.57157063)%
    \lineheight{1}%
    \setlength\tabcolsep{0pt}%
    \put(0,0){\includegraphics[width=\unitlength,page=1]{figures/assignments/supp_assignments.pdf}}%
    \put(0.05,0.55340361){\color[rgb]{0,0,0}\makebox(0,0)[lt]{\lineheight{1.25}\smash{\begin{tabular}[t]{l}Image\end{tabular}}}}%
    \put(0.26469882,0.55340361){\color[rgb]{0,0,0}\makebox(0,0)[lt]{\lineheight{1.25}\smash{\begin{tabular}[t]{l}GT\end{tabular}}}}%
    \put(0.88,0.55340361){\color[rgb]{0,0,0}\makebox(0,0)[lt]{\lineheight{1.25}\smash{\begin{tabular}[t]{l}$\mathbf{C}_{k,\cdot}$\end{tabular}}}}%
    \put(0.63,0.55340361){\color[rgb]{0,0,0}\makebox(0,0)[lt]{\lineheight{1.25}\smash{\begin{tabular}[t]{l}$\mathbf{C}_{person,\cdot}$\end{tabular}}}}%
    \put(0.48,0.55340361){\color[rgb]{0,0,0}\makebox(0,0)[lt]{\lineheight{1.25}\smash{\begin{tabular}[t]{l}$\mathbf{C}_{bg,\cdot}$\end{tabular}}}}%
    \put(0,0){\includegraphics[width=\unitlength,page=2]{figures/assignments/supp_assignments.pdf}}%
  \end{picture}%
\endgroup%

%% file: figures/iteration/lr_iters.pdf_tex
\begingroup%
  \makeatletter%
  \providecommand\color[2][]{%
    \errmessage{(Inkscape) Color is used for the text in Inkscape, but the package 'color.sty' is not loaded}%
    \renewcommand\color[2][]{}%
  }%
  \providecommand\transparent[1]{%
    \errmessage{(Inkscape) Transparency is used (non-zero) for the text in Inkscape, but the package 'transparent.sty' is not loaded}%
    \renewcommand\transparent[1]{}%
  }%
  \providecommand\rotatebox[2]{#2}%
  \newcommand*\fsize{\dimexpr\f@size pt\relax}%
  \newcommand*\lineheight[1]{\fontsize{\fsize}{#1\fsize}\selectfont}%
  \ifx\svgwidth\undefined%
    \setlength{\unitlength}{1005.99993896bp}%
    \ifx\svgscale\undefined%
      \relax%
    \else%
      \setlength{\unitlength}{\unitlength * \real{\svgscale}}%
    \fi%
  \else%
    \setlength{\unitlength}{\svgwidth}%
  \fi%
  \global\let\svgwidth\undefined%
  \global\let\svgscale\undefined%
  \makeatother%
  \begin{picture}(1,0.64119885)%
    \lineheight{1}%
    \setlength\tabcolsep{0pt}%
    \put(0,0){\includegraphics[width=\unitlength,page=1]{figures/iteration/lr_iters.pdf}}%
    \put(0,0){\includegraphics[width=\unitlength,page=2]{figures/iteration/lr_iters.pdf}}%
    \put(0.05,0.61103185){\color[rgb]{0,0,0}\makebox(0,0)[lt]{\lineheight{1.25}\smash{\begin{tabular}[t]{l}Image\end{tabular}}}}%
    \put(0.27,0.61103185){\color[rgb]{0,0,0}\makebox(0,0)[lt]{\lineheight{1.25}\smash{\begin{tabular}[t]{l}$T=1$\end{tabular}}}}%
    \put(0.48067397,0.61103185){\color[rgb]{0,0,0}\makebox(0,0)[lt]{\lineheight{1.25}\smash{\begin{tabular}[t]{l}$T=2$\end{tabular}}}}%

    \put(0.67084794,0.61103185){\color[rgb]{0,0,0}\makebox(0,0)[lt]{\lineheight{1.25}\smash{\begin{tabular}[t]{l}$T=3$\end{tabular}}}}%
    \put(0.87,0.61103185){\color[rgb]{0,0,0}\makebox(0,0)[lt]{\lineheight{1.25}\smash{\begin{tabular}[t]{l}$T=4$\end{tabular}}}}%

  \end{picture}%
\endgroup%